\documentclass[10pt,twocolumn,letterpaper]{article}

\usepackage[pagenumbers]{cvpr}

\usepackage{xspace}
\newcommand{\paper}{FAH\xspace}
\newcommand{\papershort}{FAH\xspace}

\usepackage[T1]{fontenc}
\usepackage{graphicx}
\usepackage{placeins}
\usepackage{booktabs}
\usepackage{colortbl}
\definecolor{fahrow}{RGB}{232,244,252} % Light blue for FAH benchmark rows.
\usepackage{multirow}
\usepackage{pifont}
\newcommand{\cmark}{\ding{51}}
\newcommand{\xmark}{\ding{55}}
\usepackage{tikz}
\usetikzlibrary{arrows.meta,positioning,fit,backgrounds,decorations.pathmorphing,calc}
\newcommand{\best}[1]{\textbf{#1}}

\newcommand{\mBrel}{\papershort-B\xspace}
\newcommand{\mS}{\papershort-S\xspace}
\newcommand{\mL}{\papershort-L\xspace}
\newcommand{\mH}{\papershort-H\xspace}

\newcommand{\mMHR}{MHR-ref\xspace}
\newcommand{\tabfont}{\footnotesize}
\newcommand{\oursI}{\mL}

\newcommand{\oursW}{\papershort-World\xspace}
\newcommand{\mLV}{\papershort-L video\xspace}
\newcommand{\mSV}{\papershort-S video\xspace}
\newcommand{\mBV}{\papershort-B video\xspace}
\newcommand{\mHV}{\papershort-H video\xspace}
\newcommand{\mLW}{\papershort-L world\xspace}
\renewcommand{\topfraction}{0.85}
\renewcommand{\textfraction}{0.1}
\renewcommand{\floatpagefraction}{0.85}
\tikzset{
  fsk/.style={decorate, decoration={random steps, segment length=3pt, amplitude=0.2pt, pre length=1pt, post length=2.0mm}, line cap=round, line join=round, line width=0.65pt, color=black!62},
  ffl/.style={fsk, -{Stealth[round, length=1.9mm, width=1.6mm]}},
  ffd/.style={ffl, dash pattern=on 2.2pt off 1.6pt},
  fpcap/.style={font=\scriptsize, inner sep=1pt, anchor=north, align=center, text=black!85},
  falab/.style={font=\tiny\itshape, inner sep=1pt, text=black!65},
}
\newcommand{\figtile}[6]{%
  \node[anchor=north west, minimum width=#4, minimum height=#5, inner sep=0pt] (#1) at #2 {};
  \begin{scope}
    \clip[rounded corners=3pt] (#1.north west) rectangle (#1.south east);
    \ifx#6c\node[inner sep=0pt] at (#1.center) {\includegraphics[width=#4]{#3}};%
    \else\ifx#6h\node[inner sep=0pt] at (#1.center) {\includegraphics[height=#5]{#3}};%
    \else\node[inner sep=0pt] at (#1.center) {\includegraphics[width=#4,height=#5,keepaspectratio]{#3}};\fi\fi
  \end{scope}
  \ifx#6f\draw[rounded corners=3pt, black!15, line width=0.4pt] (#1.north west) rectangle (#1.south east);\fi
}
\newcommand{\fahglyph}[2]{%
  \node[anchor=north, font=\footnotesize\bfseries, text=black!80, inner sep=1pt] (#1t) at #2 {FAH};
  \node[anchor=north, draw=black!45, line width=0.5pt, rounded corners=1.5pt, minimum width=0.42in, minimum height=0.16in, font=\tiny, text=black!70] (#1d) at ([yshift=-0.05in]#1t.south) {detector};
  \node[anchor=north, inner sep=0pt] (#1c) at ([yshift=-0.10in]#1d.south) {\includegraphics[width=0.26in]{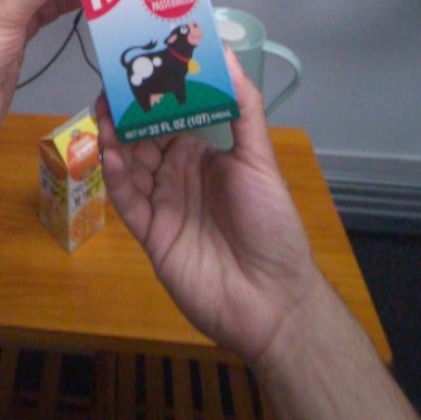}};
  \coordinate (#1e) at ([yshift=-0.10in]#1c.south);
  \draw[draw=black!50, fill=black!6, line width=0.5pt, rounded corners=1pt] ([xshift=-0.24in]#1e) -- ([xshift=0.24in]#1e) -- ([xshift=0.13in, yshift=-0.30in]#1e) -- ([xshift=-0.13in, yshift=-0.30in]#1e) -- cycle;
  \node[font=\tiny, text=black!70] at ([yshift=-0.15in]#1e) {ViT};
  \node[anchor=north, draw=black!45, fill=figmhr!10, line width=0.5pt, rounded corners=1.5pt, minimum width=0.42in, minimum height=0.16in, font=\tiny, text=black!70] (#1h) at ([yshift=-0.40in]#1e) {heads};
  \draw[ffl, line width=0.5pt] (#1d.south) -- (#1c.north);
  \draw[ffl, line width=0.5pt] (#1c.south) -- (#1e);
  \draw[ffl, line width=0.5pt] ([yshift=-0.30in]#1e) -- (#1h.north);
}

\definecolor{cvprblue}{rgb}{0.21,0.49,0.74}
\usepackage[breaklinks,colorlinks,allcolors=cvprblue]{hyperref}

\def\confName{arXiv}
\def\confYear{2026}

\title{Fast and Accurate Hand Estimation in Pixels, Interaction and World Space}

\author{Jinkun Cao\thanks{This paper is an independent personal project and not related to Jinkun Cao's position at Amazon FAR.}\\
{\tt\small jinkuncao@gmail.com}
}

\begin{document}
\maketitle
\begin{abstract}
Monocular hand mesh estimators commonly predict rig parameters and project the mesh to obtain 2D keypoints.
We find that directly predicted 3D joints agree better with 2D predictions, and fitting a mesh to predicted points reduces mesh error.
We present \paper (\textbf{F}ast and \textbf{A}ccurate \textbf{H}and Estimation), a family of models built on SAM~3D Body's hand branch through fine-tuning and feature distillation.
The models predict corresponding 2D and 3D joints and surface points, fit MANO or MHR-Hand, and optionally estimate anatomical parts and contact without object geometry at inference.
Causal refinement reduces acceleration error, and camera estimation places the hands in world space.
Experiments show improvements in image-space alignment, hand geometry, contact estimation and world-space tracking over the evaluated baselines.
The default ViT-L model runs at 65\,fps with mesh fitting on an RTX 4070 Ti SUPER, or 49\,fps with parts and contact enabled.
The smallest ViT-S model reaches 125\,fps without mesh fitting, parts or contact.
Code: \url{https://github.com/noahcao/FAH}.
\end{abstract}

\begin{figure*}[t]
  \centering
  \includegraphics[width=\linewidth]{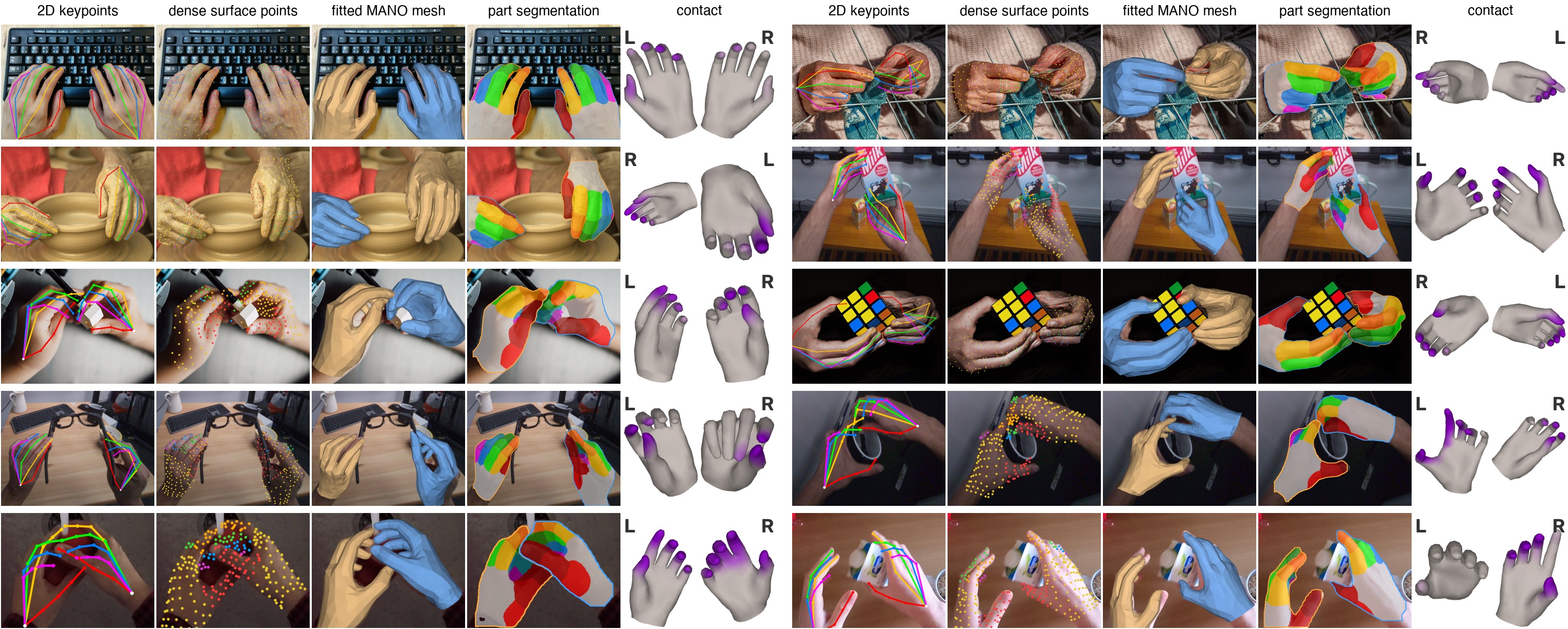}
  \caption{\textbf{Five outputs of \mH (\paper with ViT-H backbone) on ten scenes.} Left to right: 2D keypoints, visible dense points at predicted 2D locations, a MANO mesh fitted to the 3D points, optional part labels within the detector mask, and contact on the fitted mesh (gray to violet, probability 0--1; contact-side view). Outlines: right hand blue, left orange. Parts: palm gray; thumb to pinky red, orange, green, blue and magenta; darker shades are more distal. Image credits: \cref{sec:supp_credits}.}
  \label{fig:teaser}
\end{figure*}

\section{Introduction}
\label{sec:intro}
Augmented reality and robotic data pipelines from human video~\cite{anyteleop,dexcap,opentelevision} require hand estimates that align with visible fingers, recover 3D geometry and contact, remain stable over time and run in real time.
Most mesh estimators regress MANO~\cite{mano} or MHR~\cite{mhr} parameters and project the reconstructed mesh to obtain 2D keypoints~\cite{hamer,wilor,hamba,anyhand,sam3dbody,hawor}.
Errors in articulation, shape and camera placement propagate to the 2D landmarks; rig constraints and linear blend skinning can further limit pixel alignment.

Our controlled comparison shows that the output representation contributes to this error: within one network, directly predicted 3D points agree better with the 2D predictions, and fitting a mesh to them is more accurate than predicting MANO parameters (\cref{sec:m_analysis}).

\paper builds on SAM~3D Body's hand branch~\cite{sam3dbody}, replacing its decoder and rig head with a heatmap head for corresponding 2D and 3D joints and surface points.
A separate fit to MANO, MHR-Hand or both regularizes the 3D surface while preserving the direct 2D predictions.
SAM masks~\cite{sam2,sam3} supervise the surface, part labels and detector masks, which also guide fingertip alignment at inference.
Optional heads predict anatomical parts and contact without object geometry.
Causal refinement extends the model to video, and camera estimation places the hands in world space.
Training also requires consistent labels: kinematic MANO joints, mesh-regressed joints, ARKit skeletons and fingertip vertices define different target locations~\cite{freihand,ho3d,dexycb,hot3d,egodex}.
We harmonize these conventions before training.
To present \paper, we make three major contributions:

\textbf{Point prediction then mesh fitting.} We predict corresponding 2D and 3D hand joints and surface points, then fit a mesh separately. In a controlled comparison, this improves mesh accuracy over direct rig-parameter prediction.

\textbf{Contact, video and world-space estimation.} Predicted hand geometry improves contact estimation without object geometry at inference. Causal refinement reduces pose acceleration error and improves contact timing; camera estimation supports causal and offline world-space tracking.

\textbf{Hand estimation at interactive rates.} The model family runs at 27--101\,fps with mesh fitting on an RTX 4070 Ti SUPER. The smallest model also runs above 30\,fps on an M3 Pro MacBook, supporting interactive applications and large-scale video processing.

\section{Related Work}
\label{sec:related}

\paragraph{Hand mesh recovery.}
Single-image methods regress MANO~\cite{mano} parameters from a hand crop: HaMeR~\cite{hamer} trains a large ViT on a mixture of 2D and 3D data, WiLoR~\cite{wilor} adds a detector and in-the-wild data, Hamba~\cite{hamba} uses graph-guided Mamba blocks, and AnyHand~\cite{anyhand} adds synthetic data.
SAM~3D Body~\cite{sam3dbody} jointly recovers the body and hands using a promptable decoder and the MHR rig~\cite{mhr}. We build on its hand branch.
Other methods regress mesh vertices directly~\cite{meshgraphormer,simplehand}, while feedback on image features~\cite{pymaf} and silhouette supervision~\cite{wildhands} improve mesh-image alignment.
FAH predicts corresponding 2D and 3D surface points before fitting a mesh.

\paragraph{Egocentric video and world space.}
Egocentric hand reconstruction captures manipulation, while world-space estimates recover trajectories beyond the moving camera frame.
HaWoR~\cite{hawor} combines a temporal model with masked DROID-SLAM~\cite{droidslam} and metric depth~\cite{metric3dv2} to recover world-frame trajectories.
ViDiHand~\cite{wang2026vidihand} uses a video diffusion model for detector-free clip inference and introduces coverage-aware metrics.
Temporal filters~\cite{oneeuro,smoothnet} and video hand models~\cite{deformer} address motion stability.
\paper adds causal pose refinement and camera estimation to a fast per-frame model while preserving its direct 2D stream.

\paragraph{Hand contact and interaction.}
Hand-object contact identifies where the hand touches an object, providing cues about manipulation and physical interaction.
HACO~\cite{haco} estimates dense per-vertex hand contact from a single image, following body-scene contact estimators~\cite{bstro,deco}.
Contact4D~\cite{contact4d} provides synchronized whole-body and hand motion with ground-truth finger contact during dexterous operations.
Distance rules on reconstructed hand and object meshes~\cite{mow} require object shape and pose. FAH predicts contact from hand features and geometry.
Joint Diffusion~\cite{jhod} models hands and objects jointly for grasp generation, while Omnigrasp~\cite{omnigrasp} learns simulated humanoid control for grasping and transporting diverse objects.

\section{Method}
\label{sec:m_method}

\begin{figure*}[t]
\centering
\definecolor{figmhr}{rgb}{0.25,0.50,0.85}\definecolor{figmano}{rgb}{0.75,0.10,0.15}
\begin{tikzpicture}[>=Stealth, every node/.style={inner sep=0pt}]
\pgfmathsetseed{7}
\def\TW{1.06in}\def\TH{0.63in}
% hero: the input frame with the detector's outputs drawn on it
\figtile{hero}{(0,0)}{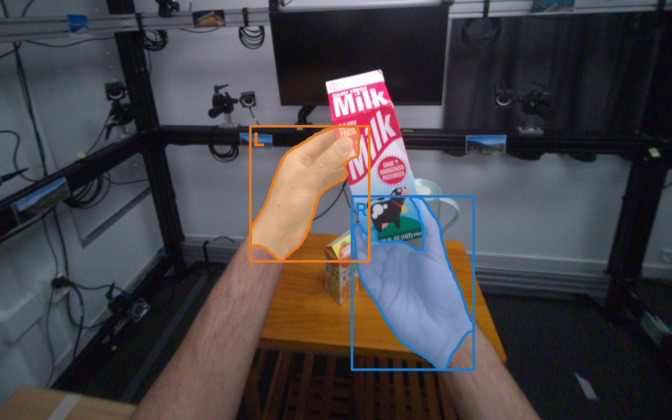}{2.30in}{1.44in}{h}
\node[fpcap] at ([yshift=-2pt]hero.south) {input frame + detector outputs};
% glyph
\fahglyph{g}{([xshift=0.86in, yshift=0.02in]hero.north east)}
% frame <-> detector: the image goes in, boxes / handedness / masks come back
\coordinate (gdw) at (gd.west);
\coordinate (hin) at ([yshift=0.05in]hero.east |- gdw);
\coordinate (hout) at ([yshift=-0.12in]hero.east |- gdw);
\draw[ffl] (hin) -- node[above, falab] {image} ([yshift=0.05in]gdw);
\draw[ffl] (gdw) .. controls +(-0.25in,0) and +(0.30in,0) .. node[pos=0.5, below, falab, align=center, yshift=-2pt, xshift=-5pt] {boxes,\\[-1pt]handedness,\\[-1pt]masks} (hout);
% mask link inside the glyph: detector -> heads (left side)
\coordinate (ms) at ([yshift=0.03in]gd.south west);
\coordinate (mv) at ([xshift=-0.08in]ms);
\fill[black!62] (ms) circle (1.1pt);
\draw[fsk, dash pattern=on 1.6pt off 1.3pt, line width=0.5pt, -{Stealth[round, length=1.5mm, width=1.3mm]}] (ms) -- (mv) -- (mv |- gh.west) -- (gh.west);
\node[falab, anchor=east] at ([xshift=-1pt]$(mv)!0.55!(mv |- gh.west)$) {mask};
% 2x2 outputs
\coordinate (G) at ([xshift=0.48in]hero.north east -| gt.east);
\figtile{pts}{(G)}{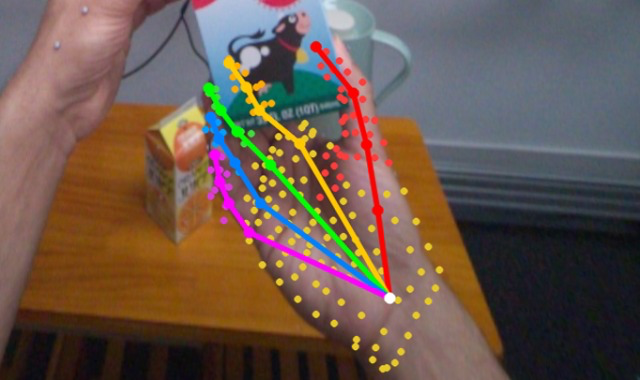}{\TW}{\TH}{c}
\node[anchor=north east, inner sep=0pt, fill=white, rounded corners=2pt] (pin) at ([xshift=-1.5pt, yshift=-1.5pt]pts.north east) {\includegraphics[height=0.28in]{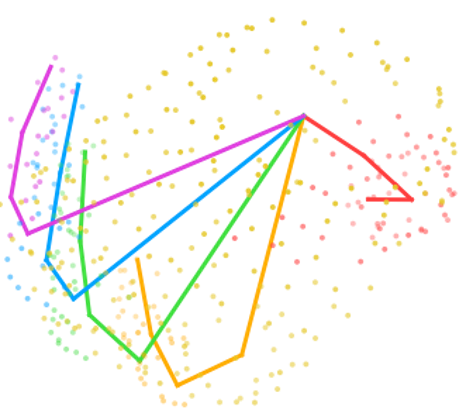}};
\draw[rounded corners=2pt, black!20, line width=0.35pt] (pin.south west) rectangle (pin.north east);
\figtile{msh}{([xshift=0.34in]pts.north east)}{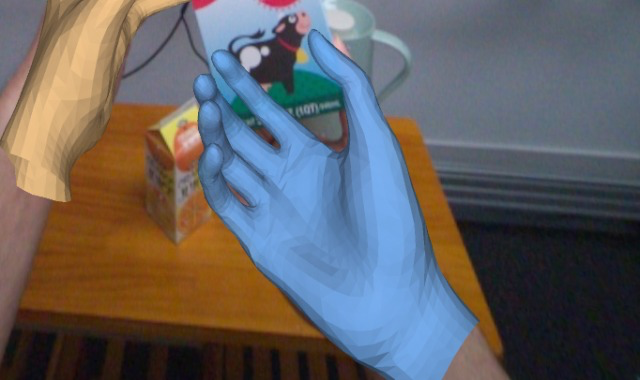}{\TW}{\TH}{c}
\node[anchor=north east, inner sep=0pt, fill=white] (min) at ([xshift=-1.5pt, yshift=-1.5pt]msh.north east) {\includegraphics[height=0.28in]{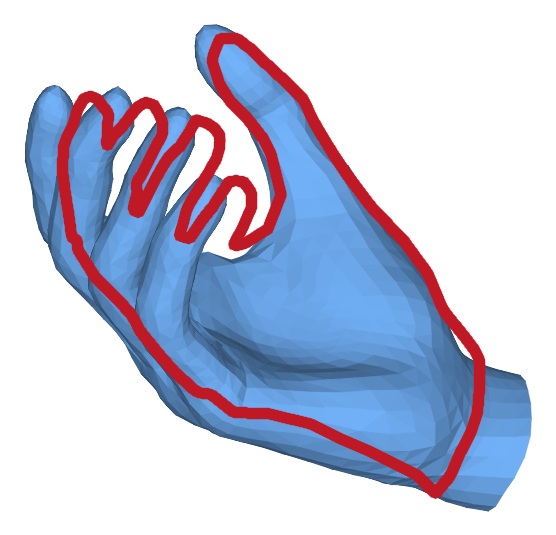}};
\draw[rounded corners=2pt, black!20, line width=0.35pt] (min.south west) rectangle (min.north east);
\figtile{prt}{([yshift=-0.27in]pts.south west)}{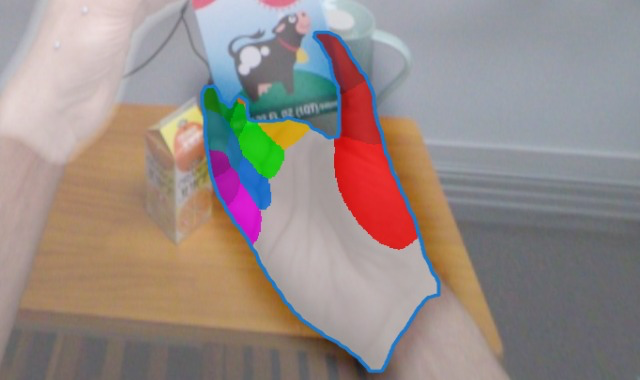}{\TW}{\TH}{c}
\figtile{ctc}{([yshift=-0.27in]msh.south west)}{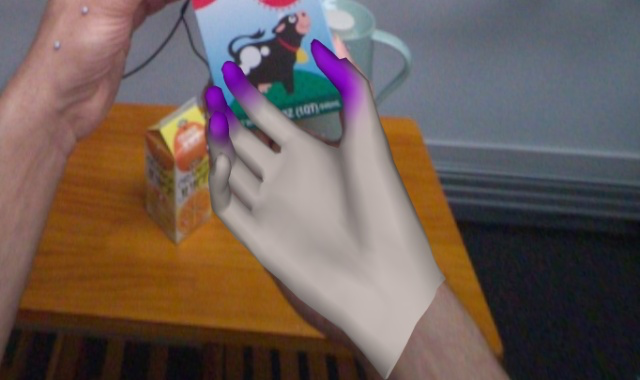}{\TW}{\TH}{c}
\node[anchor=north east, inner sep=0pt, fill=white] (cin) at ([xshift=-1.5pt, yshift=-1.5pt]ctc.north east) {\includegraphics[height=0.28in]{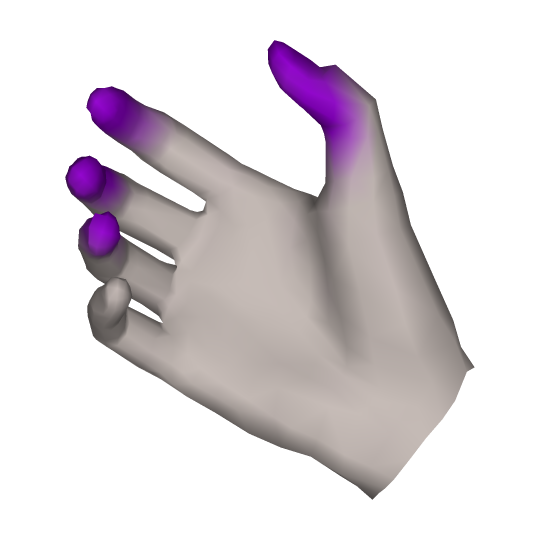}};
\draw[rounded corners=2pt, black!20, line width=0.35pt] (cin.south west) rectangle (cin.north east);
\node[fpcap] (lpts) at ([yshift=-2pt]pts.south) {2D + 3D hand points};
\node[fpcap] at ([yshift=-2pt]msh.south) {\textcolor{figmhr}{MHR-Hand} / \textcolor{figmano}{MANO}};
\node[fpcap] at ([yshift=-2pt]prt.south) {parts in the mask};
\node[fpcap] at ([yshift=-2pt]ctc.south) {contact on the surface};
% heads -> outputs
\coordinate (ghe) at (gh.east);
\draw[ffl] ([yshift=0.03in]ghe) .. controls +(0.18in,0.05in) and +(-0.20in,0) .. ([yshift=-0.10in]pts.west);
\node[falab, anchor=south east] at ([xshift=-4pt, yshift=-0.06in]pts.west) {heatmap head};
\draw[ffl] (ghe) .. controls +(0.18in,0) and +(-0.18in,0) .. node[pos=0.55, below, falab, yshift=-1pt] {part head} ([yshift=0.05in]prt.west);
\coordinate (gapy) at ($(lpts.south)!0.5!(prt.north)$);
\coordinate (cw1) at ([xshift=-0.02in]gapy -| prt.east);
\coordinate (cen) at ([yshift=-0.12in]ctc.north west);
\draw[ffl] ([yshift=-0.03in]ghe) .. controls +(0.10in,0) and +(-0.25in,0) .. ([xshift=0.06in]gapy -| prt.west) -- (cw1) .. controls +(0.17in,0) and +(-0.17in,0) .. (cen);
\node[falab, anchor=south] at ([xshift=0.12in, yshift=0.5pt]cw1) {contact head};
\draw[ffd] (pts.east) -- node[above, falab] {fit} (msh.west);
% over time / world strip
\coordinate (S) at ([yshift=-0.25in]hero.south west |- prt.south);
\def\SW{0.70in}\def\SH{0.54in}
\figtile{s1}{(S)}{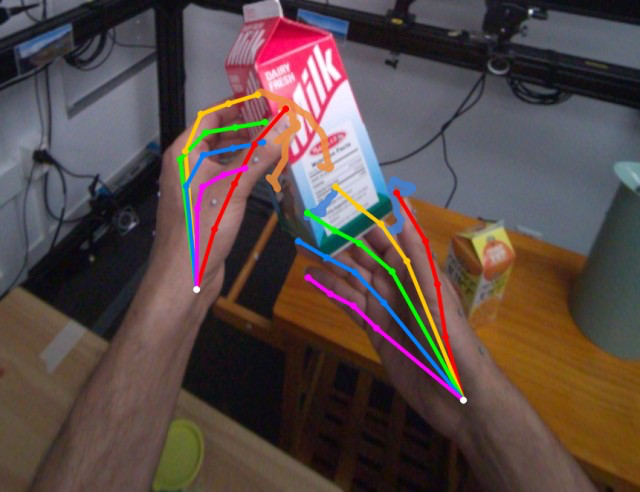}{\SW}{\SH}{c}
\figtile{s2}{([xshift=0.04in]s1.north east)}{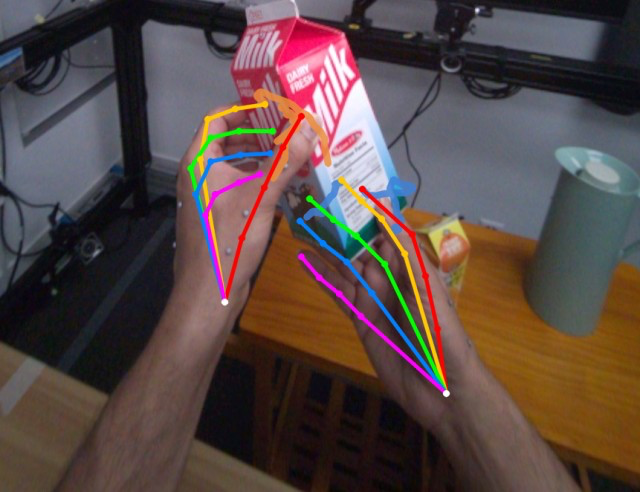}{\SW}{\SH}{c}
\figtile{s3}{([xshift=0.04in]s2.north east)}{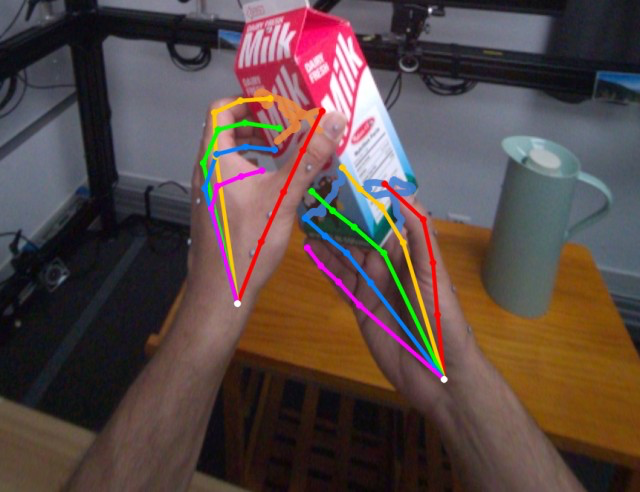}{\SW}{\SH}{c}
\figtile{s4}{([xshift=0.04in]s3.north east)}{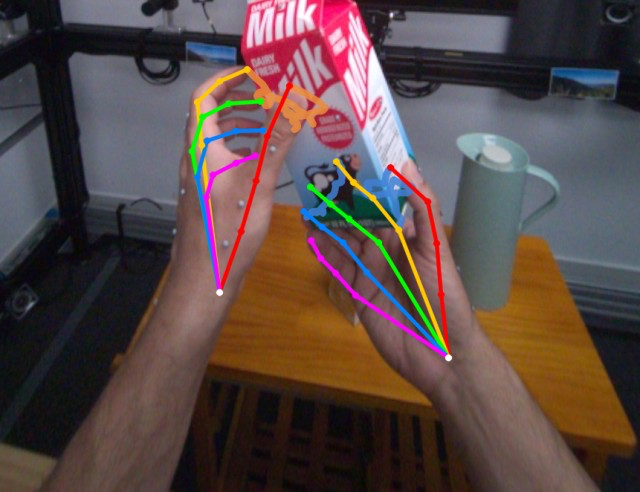}{\SW}{\SH}{c}
\figtile{s5}{([xshift=0.04in]s4.north east)}{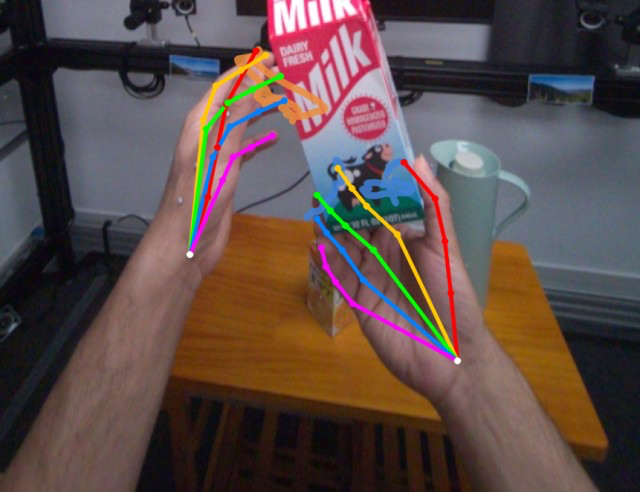}{\SW}{\SH}{c}
\figtile{s6}{([xshift=0.04in]s5.north east)}{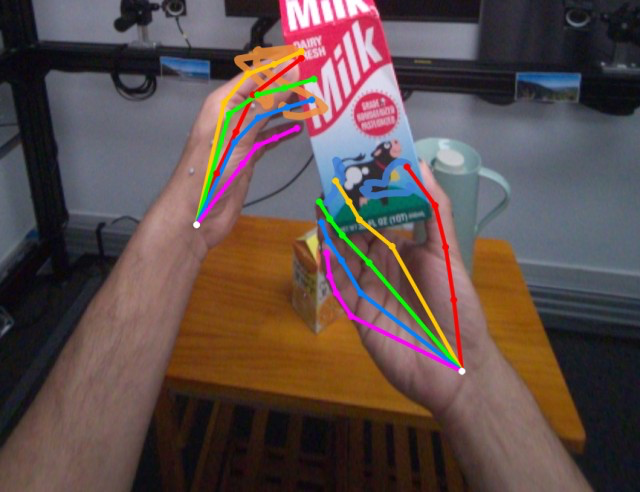}{\SW}{\SH}{c}
\figtile{s7}{([xshift=0.04in]s6.north east)}{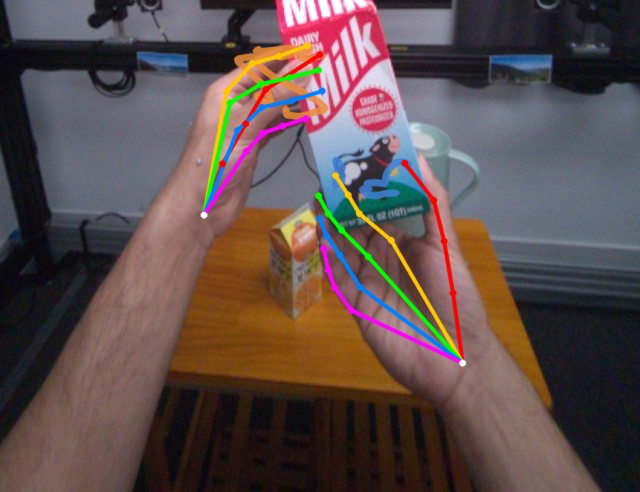}{\SW}{\SH}{c}
\coordinate (wx) at ([xshift=-0.70in]ctc.east);
\figtile{wld}{(s7.north -| wx)}{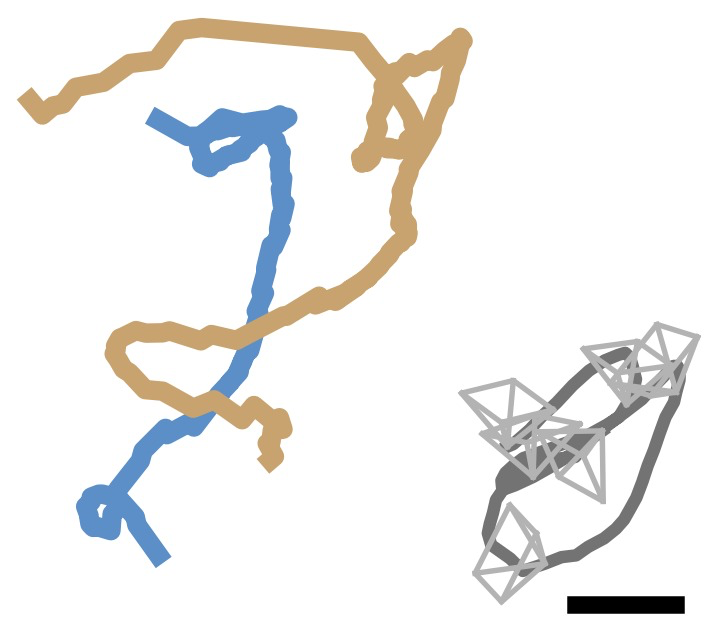}{0.70in}{\SH}{f}
\node[fpcap] at ([yshift=-2pt]$(s1.south west)!0.5!(s7.south east)$) {causal video track: refined joints with fingertip trails, $t{-}24 \ldots t$};
\node[anchor=east, font=\tiny, inner sep=1pt] at ([xshift=0.52in, yshift=-0.52in]wld.north west) {10\,cm};
\node[fpcap] at ([yshift=-2pt]wld.south) {world space};
\draw[ffd] (s7.east) -- node[above, falab, align=center] {camera\\[-1pt]model} (wld.west);
\node[falab, anchor=south west] at ([yshift=2pt]s1.north west) {video model};
\end{tikzpicture}
\caption{\textbf{FAH on a training-disjoint HOT3D frame.} The detector supplies boxes, handedness, masks and crops. The heatmap head predicts 2D and root-relative 3D joints and surface points (inset). Dashed arrows: MHR-Hand and MANO fitting; inset: side view, MANO outlined. The part head labels mask pixels; contact is shown on fitted MANO (gray to violet, probability 0--1; inset: palm side). Bottom: causal joint refinement, frames four apart with thumb/index fingertip trails, followed by world-space placement (camera gray, wrists colored). Blue: right hand; orange: left.}
\label{fig:pipeline}
\end{figure*}

\Cref{fig:pipeline} shows the prediction paths. \paper builds on existing architectures, using SAM~\cite{sam2,sam3} masks and SAM~3D Body~\cite{sam3dbody} supervision. Separate training stages exploit datasets with different annotations (\cref{sec:supp_common}).

\subsection{Hand detection}
\label{sec:m_det}
A one-stage YOLOv8m segmentation network~\cite{yolov8} predicts hand boxes, handedness and instance masks at a 640\,px input in FP16.
The detector trains on about 96k quality-checked images, including two-hand frames from HOT3D and OakInk2 training sequences (\cref{sec:supp_det}).
Target boxes enclose both SAM~3~\cite{sam3} hand masks and visible keypoints.
The pose network re-scores each box with \emph{presence} and \emph{handedness-consistency} logits and re-runs doubtful sides mirrored; identity NMS suppresses duplicate hypotheses using predicted wrist distances, and hands touching the crop border are re-cropped once.

\subsection{Pose network}
\label{sec:m_pose}
\paragraph{Teacher and student.}
We extract SAM~3D Body's hand branch (DINOv3-H+ backbone, promptable decoder and MHR head) and fine-tune it with the benchmark sequences held out.
Feature distillation initializes DINOv3 ViT-S/16, ViT-B/16 and ViT-L/16 students, which then train with the supervision below.
\mL uses ViT-L/16 on 320\,px crops; \mS and \mBrel use the smaller students, and \mH retains the teacher's ViT-H+/16 at 448\,px.
\paragraph{Decoder-free dense head.}
The pose network uses a heatmap head directly on the encoder tokens, without the teacher's promptable decoder or rig head.
It predicts 21 MANO-convention joints, two points inside each of the 20 bones and 400 farthest-point samples of the 778 MANO vertices, each with crop-normalized 2D (soft-argmax) and root-relative 3D coordinates.
A root-depth output predicts $\log(z\,s/f)$ for a crop of size $s$ and focal length $f$, and the $x, y$ translation follows in closed form from the 2D and 3D points at that depth.
The full 778-vertex surface is reconstructed from the 400 points through the lowest 200 eigenfunctions of the template's cotangent Laplacian.
\paragraph{Supervision.}
Joints and bone points use ground-truth labels where available, with teacher targets on sources lacking 3D labels.
Surface points use ground-truth meshes or SAM~2.1 pseudo-masks with a weak teacher-mesh prior.
A reprojection-consistency loss supervises projected 3D points on 2D-labeled data, and incorrectly mirrored crops train the verification logits.
Let $\hat{\mathbf u}_i$ and $\hat{\mathbf X}_i$ denote crop-normalized 2D and root-relative 3D predictions, with targets $\mathbf u_i^*$ and $\mathbf X_i^*$. The point-supervision terms have the form
\begin{equation}
\label{eq:point_loss}
\begin{aligned}
\mathcal L_{\mathrm{pts}} ={}&
 \sum_i a_i^{2D}\rho_{0.02}(\hat{\mathbf u}_i-\mathbf u_i^*)\\
 &+\lambda_{3D}\sum_i a_i^{3D}\rho_{0.01}(\hat{\mathbf X}_i-\mathbf X_i^*).
\end{aligned}
\end{equation}
$\rho_\beta$ sums coordinate-wise smooth-L1 penalties; 3D coordinates are in meters. Weights $a_i^{2D},a_i^{3D}$ encode validity, confidence and source normalization. Ground-truth and teacher terms are accumulated separately; joint and bone losses use $\lambda_{3D}=3$ and teacher weight $0.5$. Root-depth, visibility, mask and verification losses complete the supervision (\cref{sec:supp_dense}).
For reprojection, let $\pi_{K,A}$ project camera-space points through intrinsics $K$ and the crop transform $A$, returning crop-normalized coordinates. We minimize
\begin{equation}
\label{eq:reprojection_loss}
\mathcal L_{\mathrm{rep}}=
 \sum_i r_i\rho_{0.02}\!\left(
 \pi_{K,A}(\hat{\mathbf X}_i+\bar{\mathbf t})-\bar{\mathbf u}_i\right).
\end{equation}
Joint targets $\bar{\mathbf u}_i$ use 2D labels where available and detached direct predictions otherwise. Translation $\bar{\mathbf t}$ is fitted at ground-truth or teacher depth and held fixed during differentiation. Visible surface targets are detached direct 2D predictions; weights $r_i$ mask invalid targets and downweight unlabeled joints.

\paragraph{Label conventions.}
We harmonize 3D sources to kinematic MANO joints and correct per-dataset 2D offsets.
For EgoDex's ARKit skeleton~\cite{egodex}, the conversion uses a reference model trained without EgoDex (\cref{sec:labelconv}).

\subsection{Mesh fitter}
\label{sec:m_mesh}
The mesh stage fits MANO~\cite{mano}, MHR-Hand (the hand rig of MHR~\cite{mhr}, as used by SAM~3D Body), or both, returning parameters and a mesh for each. 
MANO fitting has three steps.
First, Horn's quaternion method~\cite{horn1987} aligns each template part to the corresponding predicted points.
A learned fitter then predicts MANO pose and shape from this initialization and the points.
One Gauss--Newton step refines the parameters against the predicted geometry, expressed relative to the predicted wrist so that the central-difference Jacobian is stable; it is evaluated in one batched forward pass.
The fit runs as one CUDA graph.
MANO-convention fingertips can stop short of the visible finger end.
Where the detector mask resolves a fingertip, we extend its 2D prediction along the distal segment to the mask boundary, by at most 0.6 segment lengths.
The mesh's distal phalanges stretch accordingly; the benchmark 3D joints retain the MANO convention (\cref{sec:m_analysis,sec:supp_align}).
The MHR-Hand fit uses the same points with its own learned initializer and Gauss--Newton step, at comparable throughput (\cref{sec:supp_mesh}).

\subsection{Parts and contact on the hand surface}
\label{sec:m_contact}
\paragraph{Part segmentation.}
The optional part head predicts background, palm and three segments per finger using its own queries on the shared features and image-guided refinement at half the crop resolution.
It trains with the pose network frozen, using cross-entropy and a loss that matches the part union to the hand mask.
Reference labels are MANO part renders clipped to quality-checked SAM~2.1 masks, from ground-truth meshes where available and our mesh on 2D-only data (\cref{sec:supp_parts}).
For pixel $x$, let $p_{x,c}$ be the predicted class probability, $y_x$ its part label and $q_x=1-p_{x,0}$ the foreground union. With hand-mask target $M$, the objective is
\begin{equation}
\label{eq:part_loss}
\begin{aligned}
\mathcal L_{\mathrm{part}}={}&
 -\frac{\sum_{x\in\Omega}w_{y_x}\log p_{x,y_x}}
 {\sum_{x\in\Omega}w_{y_x}}\\
 &+\lambda_{\mathrm{union}}\bigl[
 \mathrm{BCE}(q,M)+\lambda_D\mathcal D(q,M)\bigr],\\
\mathcal D(q,M)={}&1-
 \frac{2\sum_x q_xM_x+1}{\sum_xq_x+\sum_xM_x+1}.
\end{aligned}
\end{equation}
$\Omega$ contains labeled pixels; $w_c=3$ for finger classes and $1$ otherwise. Union BCE uses crops with a valid hand mask and negative crops (zero target); Dice is averaged only over hand-mask crops. Terms without valid labels are omitted. The pose network remains frozen.

At inference, each detector-mask pixel takes its most likely foreground part; the part union therefore equals the detector mask. This head needs no mesh.
\paragraph{Contact.}
A contact head on the frozen pose network estimates, for each of the 400 surface points, the probability that it touches an object.
Each point query combines heatmap-pooled image features, predicted root-relative 3D position and visibility, and a learned point embedding.
Together with the hand's pose token, four transformer layers (256-d) attend over the points and cross-attend to pooled heatmap features and contact-branch tokens.
The contact branch is a trainable copy of the encoder's last four ViT blocks, operating on frozen encoder features; pose, points and parts are unchanged.
At inference, the mesh's spectral basis maps per-point probabilities to the 778 MANO vertices, clipped to $[0,1]$.
Finger and palm summaries take the maximum vertex probability, and a separate hand-level output flags contact.
Training labels mark MANO vertices within 5, 10, 20 and 35\,mm of or inside a ground-truth object mesh on HOT3D, DexYCB, HO-3D, OakInk2, egocentric ARCTIC~\cite{arctic} and HOI4D~\cite{hoi4d} training sequences, with InterHand2.6M hands as contact-free examples (699k crops, 102k of them from ARCTIC and HOI4D); MOW~\cite{mow} supervises only the 35\,mm output.
All four sizes use 24k steps of class-balanced binary cross-entropy for the head and its branch.
For tolerance $k\in\{5,10,20,35\}$\,mm, let $c_{ik}$ and $y_{ik}\in\{0,1\}$ be the predicted point-contact probability and label. Over valid point labels in a minibatch, the balanced objective is
\begin{equation}
\label{eq:contact_loss}
\begin{aligned}
\mathcal B_k={}&-\frac{\sum_{i\in\mathcal V_k}y_{ik}\log c_{ik}}
 {2\max(N_k^+,1)}\\
 &-\frac{\sum_{i\in\mathcal V_k}(1-y_{ik})\log(1-c_{ik})}
 {2\max(N_k^-,1)},\\
\mathcal L_{\mathrm{contact}}={}&\frac14\sum_k
 \left[\mathcal B_k+\tfrac12\mathrm{BCE}(h_k,y_k^h)\right].
\end{aligned}
\end{equation}
$\mathcal V_k$ contains the valid labels at tolerance $k$, with positive and negative counts $N_k^+,N_k^-$. The hand-level probability $h_k$ uses the corresponding binary target $y_k^h$, averaged over labeled hands. Missing tolerances contribute zero, so MOW supervises only the 35\,mm terms.

A validation-fitted logit offset maps the selected 10\,mm threshold to 0.5.
No object geometry is used at inference.
In video, a causal 4-layer transformer over the last 32 frames of each track reads the per-frame contact logits, hand logit, joint motion and hand token and predicts residuals to them; its map is blended with the per-frame map (weight 0.25) and its hand flag uses hysteresis.
It trains on out-of-fold predictions: each of two image heads predicts the recordings excluded from its own training fold (\cref{sec:supp_chead}).

\subsection{Video model}
\label{sec:m_ref}
The video model smooths hand trajectories with causal refinement.
Per-frame outputs are linked into one track per handedness, and a 4-layer, 256-d causal transformer over the last 32 frames of a track predicts residuals to the root-relative joints and the root, and complete joint and root predictions on frames without a detection.
It trains on a ViT-B image model's predictions on HOT3D and DexYCB training clips at strides 1--3, so it also sees fast motion (\cref{sec:supp_refiner}).
At inference we blend refined and per-frame articulation with weights 0.75 and 0.25 and apply a causal One Euro filter~\cite{oneeuro}.
The refined root replaces the per-frame estimate unless it would move the hand by more than 5\,cm (\cref{sec:supp_refiner}).
The direct 2D outputs use their own speed-adaptive filter, while a handedness memory suppresses side flips.
Track proposals cover per-frame misses: a confirmed track without a confident detector box receives a box at its constant-velocity prediction, which the presence logit must accept (\cref{sec:supp_refiner}).

\subsection{Camera model}
\label{sec:m_apps}
\label{sec:m_cam}
Our causal camera model, \paper-camera, estimates each frame's pose on arrival from past and current frames only, so it runs online alongside hand estimation.
It combines DPVO~\cite{dpvo} odometry on hand-masked frames with a running median of background MoGe-2~\cite{moge2} depth estimates for metric scale.
A learned motion head uses frozen RAFT-small~\cite{raft} flow and DINOv3 ViT-S tokens to predict relative camera motion and its uncertainty.
It corrects scale where odometry sees little parallax and supplies motion where odometry fails or the rotation or translation estimates disagree beyond the predicted uncertainty, for example when hands fill the frame and odometry tracks hand edges and moving objects (\cref{sec:supp_camera}).
\oursW (``\mLW'' in tables) places the video model's hands in world space with these poses.
An offline mode keeps the trajectory of DROID-SLAM~\cite{droidslam}, masked with our hands, and fuses its per-clip metric scale with \paper-camera's.

\subsection{Efficient inference}
\label{sec:m_speed}
The GPU pipeline shares one frame upload across detection and cropping, uses FP16 TensorRT detector and encoder engines, and captures the encoder, heads and mesh stage as CUDA graphs.
Pose heads use TF32 and the numerically sensitive mesh stage fp32.
Unless stated otherwise, fps include detection, pose, the mesh fit and the fingertip step, and exclude the optional part and contact heads.

\section{Experiments}
\label{sec:m_exp}

\paragraph{Benchmarks and metrics.}
We compare \paper with the relevant baselines in five settings:
\begin{itemize}
    \item \textbf{Egocentric 3D:} HOT3D v1 (98 five-second clips from HaWoR's validation split of HOT3D's train\_aria recordings~\cite{hot3d}, excluded from training) and EgoStandard~\cite{egostandard} (120 thirty-second test clips; aggregate metrics only). Metrics cover articulation (PA-MPJPE), mesh error (PA-MPVPE), placement (root error, MRRPE) and temporal quality (acceleration error, jitter). PA-MPJPE-p~\cite{wang2026vidihand} also penalizes missed hands (\cref{sec:benchmarks}).
    \item \textbf{Image-space pose:} bench2d v2 (8.6k hands from HOT3D, CMU Panoptic, COCO-WholeBody and HInt~\cite{hamer}) and three small, model-assisted stress tests for challenging poses, low light and shadows (\cref{sec:benchmarks}). We report matched-hand PCK@0.1 and PCK with misses counted as failures (``incl.'').
    \item \textbf{Contact:} pooled vertex F1/AP and hand-level F1 on HOT3D and DexYCB s0 test, with additional evaluation on HOI4D and ARCTIC. Labels use ground-truth hand and object meshes; intervals are 95\% paired sequence-bootstrap intervals.
    \item \textbf{World space:} HOT3D v1 and EgoStandard, using W-MPJPE100, WA-MPJPE100 and RTE following WHAM~\cite{shin2024wham} and SLAHMR~\cite{slahmr} (\cref{sec:benchmarks}).
    \item \textbf{Throughput:} fps is the inverse of median end-to-end frame time on an RTX 4070 Ti SUPER, using the same pre-decoded streams (\cref{sec:m_speed,sec:supp_timing}).
\end{itemize}
Tables use \mS, \mBrel, \mL (default) and \mH; video rows add causal refinement (\mLV for \mL).

\begin{table*}[t]
\centering
\caption{\textbf{3D accuracy on HOT3D v1 and EgoStandard} (mm; accel and jitter in mm/frame$^2$; PA-p: coverage-aware PA-MPJPE~\cite{wang2026vidihand}). fps on HOT3D frames on an NVIDIA RTX 4070 Ti SUPER (16\,GB) desktop GPU under Windows~11 + WSL2, slower of two rounds (MediaPipe: GPU delegate unavailable under WSL2 / Windows); $^\S$our CUDA-graph implementation; crop-based baselines use WiLoR's detector. \mS, \mBrel, \mL and \mH (ViT-S, ViT-B, ViT-L, ViT-H+ at 448\,px) share data, losses and schedule and differ only in encoder initialization and input size (\cref{sec:supp_capacity}); their fps is end to end (detection, pose, mesh fit, fingertips; optional parts excluded), all four timed in one session. Evaluation scope: \cref{sec:benchmarks,sec:supp_chead,sec:supp_camera}; full tables: \cref{sec:supp_tables}.}
\label{tab:m3d}
\tabfont
\setlength{\tabcolsep}{2.6pt}
\begin{tabular}{lrrrrrrrrrrrrrr}
\toprule
 & & \multicolumn{7}{c}{HOT3D v1} & \multicolumn{6}{c}{EgoStandard} \\
\cmidrule(lr){3-9} \cmidrule(lr){10-15}
method & fps & PA & PA-p & PVE & root & MRRPE & accel & jitter & PA & PA-p & PVE & root & accel & jitter \\
\midrule
\multicolumn{15}{l}{\emph{per frame}} \\
MediaPipe Hands~\cite{mediapipehands} & --- & 19.61 & 67.01 & --- & 87.72 & 119.79 & 20.59 & 20.61 & --- & --- & --- & --- & --- & --- \\
HaMeR~\cite{hamer} & 4.6 & 9.46 & 14.48 & 9.24 & 41.62 & 49.26 & 6.75 & 6.84 & --- & --- & --- & --- & --- & --- \\
Hamba~\cite{hamba} & 5.0 & 9.28 & 14.31 & 9.02 & 40.66 & 47.83 & 7.22 & 7.31 & --- & --- & --- & --- & --- & --- \\
WiLoR~\cite{wilor} & 4.7 & 6.72 & 11.85 & 6.49 & 43.89 & 48.57 & 8.30 & 8.48 & 7.26 & 8.59 & 7.52 & 64.18 & 6.64 & 6.71 \\
SAM~3D Body hand~\cite{sam3dbody} & 10$^\S$ & 8.30 & --- & 8.49 & 40.00 & 52.36 & 5.62 & 5.81 & --- & --- & --- & --- & --- & --- \\
\rowcolor{fahrow} \mS & \best{101} & 5.25 & 5.51 & 4.89 & 11.94 & 16.20 & 5.60 & 5.87 & 7.31 & 7.60 & 7.28 & 30.66 & 5.55 & 5.61 \\
\rowcolor{fahrow} \mBrel & 89 & 4.57 & 4.84 & 4.28 & 9.76 & 13.19 & 4.58 & 4.90 & 7.11 & 7.40 & \best{7.12} & 30.18 & 4.77 & 4.84 \\
\rowcolor{fahrow} \mL & 65 & 4.31 & 4.58 & 4.02 & 8.89 & 12.13 & 3.83 & 4.19 & \best{7.10} & 7.39 & \best{7.12} & 30.21 & 3.99 & 4.06 \\
\rowcolor{fahrow} \mH & 27 & 4.12 & 4.39 & 3.85 & 9.12 & 11.65 & 3.95 & 4.31 & 7.14 & 7.44 & 7.15 & \best{28.53} & 3.95 & 4.01 \\
\midrule
\multicolumn{15}{l}{\emph{video}} \\
HaWoR~\cite{hawor} & 11 & 5.89 & 14.78 & 5.57 & 29.51 & 27.67 & 1.94 & 2.03 & --- & --- & --- & --- & --- & --- \\
\rowcolor{fahrow} \mSV & 88 & 5.08 & 5.28 & 4.80 & 11.59 & 15.70 & 2.24 & 2.45 & 7.35 & 7.50 & 7.30 & 30.75 & 2.14 & 2.14 \\
\rowcolor{fahrow} \mBV & 79 & 4.42 & 4.65 & 4.19 & 9.42 & 12.69 & 1.89 & 2.15 & 7.16 & \best{7.30} & 7.15 & 30.25 & 1.91 & 1.91 \\
\rowcolor{fahrow} \mLV & 59 & 4.16 & 4.38 & 3.95 & 8.55 & 11.58 & \best{1.65} & \best{1.92} & 7.17 & 7.31 & 7.15 & 30.29 & 1.67 & 1.66 \\
\rowcolor{fahrow} \mHV & 26 & \best{3.98} & \best{4.19} & \best{3.77} & \best{8.34} & \best{10.72} & 1.66 & 1.94 & 7.20 & 7.35 & 7.18 & 28.84 & \best{1.65} & \best{1.63} \\
\bottomrule
\end{tabular}
\end{table*}

\begin{table*}[t]
\centering
\caption{\textbf{2D accuracy on bench2d v2 and the challenging-pose, low-light and shadow sets.} PCK@0.1 on matched hands; ``incl.'': missed hands count as failures; ``3D fit'': 3D joints projected at the best-fitting translation; ``overlap'': PCK@0.1 incl.\ misses on 387 overlapping hands; shadows: the 278 scored hands of shadow\_v1. Full metric set: \cref{sec:supp_tables}; fps as in \cref{tab:m3d}.}
\label{tab:m2d}
\tabfont
\setlength{\tabcolsep}{2.4pt}
\begin{tabular}{lrrrrrrrrrrrrrr}
\toprule
 & & \multicolumn{4}{c}{bench2d v2} & \multicolumn{3}{c}{challenging poses} & \multicolumn{3}{c}{low light} & \multicolumn{3}{c}{shadows} \\
\cmidrule(lr){3-6} \cmidrule(lr){7-9} \cmidrule(lr){10-12} \cmidrule(lr){13-15}
method & fps & PCK & incl. & 3D fit & overlap & recall & PCK & incl. & recall & PCK & incl. & recall & PCK & incl. \\
\midrule
MediaPipe Hands~\cite{mediapipehands} & --- & 0.65 & 0.34 & 0.55 & 0.28 & 0.05 & 0.94 & 0.05 & 0.33 & 0.65 & 0.22 & 0.58 & 0.89 & 0.51 \\
RTMPose-m~\cite{rtmpose} & 89 & 0.71 & 0.40 & --- & 0.23 & 0.07 & 0.82 & 0.06 & 0.35 & 0.70 & 0.24 & 0.45 & 0.94 & 0.42 \\
RTMW-l~\cite{rtmw} & 32 & 0.68 & 0.65 & --- & 0.56 & 0.13 & 0.84 & 0.11 & 0.85 & 0.65 & 0.56 & 0.98 & 0.89 & 0.87 \\
HaMeR~\cite{hamer} & 4.6 & 0.71 & 0.66 & 0.65 & 0.55 & 0.84 & 0.93 & 0.78 & 0.82 & 0.70 & 0.57 & 0.95 & 0.97 & 0.92 \\
Hamba~\cite{hamba} & 5.0 & 0.72 & 0.66 & 0.65 & 0.55 & 0.84 & 0.93 & 0.78 & 0.81 & 0.70 & 0.57 & 0.95 & 0.97 & 0.92 \\
WiLoR~\cite{wilor} & 4.7 & 0.69 & 0.64 & 0.77 & 0.52 & 0.84 & 0.94 & 0.78 & 0.81 & 0.67 & 0.55 & 0.95 & 0.97 & 0.92 \\
\rowcolor{fahrow} \mS & \best{101} & 0.80 & 0.78 & 0.79 & 0.60 & \best{0.90} & \best{0.99} & 0.89 & 0.96 & 0.76 & 0.73 & \best{1.00} & 0.98 & 0.97 \\
\rowcolor{fahrow} \mBrel & 89 & 0.82 & 0.80 & 0.82 & 0.62 & \best{0.90} & \best{0.99} & 0.89 & 0.96 & 0.79 & 0.76 & \best{1.00} & \best{0.99} & \best{0.99} \\
\rowcolor{fahrow} \mL & 65 & \best{0.83} & \best{0.81} & \best{0.83} & \best{0.64} & \best{0.90} & \best{0.99} & \best{0.90} & 0.96 & \best{0.81} & \best{0.78} & \best{1.00} & \best{0.99} & \best{0.99} \\
\rowcolor{fahrow} \mH & 27 & \best{0.83} & \best{0.81} & \best{0.83} & \best{0.64} & \best{0.90} & \best{0.99} & \best{0.90} & \best{0.97} & \best{0.81} & \best{0.78} & \best{1.00} & \best{0.99} & 0.98 \\
\bottomrule
\end{tabular}
\end{table*}

\subsection{Hand pose and mesh prediction}
\label{sec:m_image}
\label{sec:m_video}
\paragraph{Images.}
\oursI improves HOT3D articulation and mesh accuracy over the evaluated per-frame baselines (\cref{tab:m3d}).
On EgoStandard, the largest improvement over WiLoR is placement: root error falls from 64.18 to 30.21\,mm, with lower articulation and mesh errors.

In 2D, \mL reaches 0.83 PCK@0.1 against 0.72 for the strongest baseline; it is 6.5$\times$ faster than SAM~3D Body's hand branch (65 vs 10\,fps) and more accurate on these test hands (0.83 vs 0.61, \cref{tab:m2d,fig:mcompare}).
It leads every subset except COCO (0.74 vs 0.75), and improves accuracy including misses on overlapping hands (0.64 vs 0.56 for RTMW-l).
Recall on the three stress tests is 0.90, 0.96 and 1.00, against at most 0.84, 0.85 and 0.98 for the baselines (RTMW-l on low light and shadows).

Adding parts and then contact lowers \mL throughput from 65 to 54 and 49\,fps.
Across the released family, \mH has the lowest per-frame HOT3D articulation error at 27\,fps and the lowest EgoStandard root error (28.53\,mm), but not a lower EgoStandard PA-MPJPE (7.14 vs 7.10\,mm for \mL).

\paragraph{Video.}
The main gain from video refinement is temporal accuracy: on HOT3D, acceleration error falls from 3.83 to 1.65 (HaWoR, the evaluated video baseline: 1.94) and PA-MPJPE from 4.31 to 4.16\,mm (HaWoR: 5.89; \cref{tab:m3d}).
All four sizes show this trade-off: refinement lowers acceleration error on both datasets and HOT3D PA-MPJPE, but slightly raises EgoStandard PA-MPJPE.
On EgoStandard, acceleration error falls from 3.99 to 1.67, while PA-MPJPE changes from 7.10 to 7.17\,mm and root error from 30.21 to 30.29\,mm.

\begin{table*}[t]
\centering
\caption{\textbf{Hands in world space} (W / WA-MPJPE in mm, RTE in \%; chunks of 100 frames; mean of 3 runs, SDs in \cref{tab:supp_worldsd}; each method's own true positives). HOT3D scores guarded camera choices; earlier configurations were also scored on EgoStandard test (\cref{sec:supp_camera}). DROID-SLAM: HaWoR's camera stage (DROID-SLAM masked with HaWoR's hands, Metric3D scale, no infiller), one trajectory per clip shared by WiLoR, HaWoR and \mLV; with \mLW, our offline mode: DROID-SLAM masked with our hands, scale fused with \paper-camera's. \paper-camera: our causal camera model (\cref{sec:m_cam}). causal: every stage uses only past and current frames. fps: all stages, desktop RTX 4070 Ti SUPER, 4 EgoStandard test clips, slower of 2 rounds.}
\label{tab:mworld}
\tabfont
\setlength{\tabcolsep}{2.2pt}
\begin{tabular}{llcrrrrr@{\hspace{10pt}}rrrrrr}
\toprule
 & & & \multicolumn{5}{c}{HOT3D v1 (5\,s clips)} & \multicolumn{5}{c}{EgoStandard (30\,s clips)} & \\
\cmidrule(lr){4-8} \cmidrule(lr){9-13}
 & & & \multicolumn{2}{c}{chunks} & \multicolumn{2}{c}{whole tracks} & & \multicolumn{2}{c}{chunks} & \multicolumn{2}{c}{whole tracks} & & \\
\cmidrule(lr){4-5} \cmidrule(lr){6-7} \cmidrule(lr){9-10} \cmidrule(lr){11-12}
hands & camera & causal & W & WA & W & WA & RTE & W & WA & W & WA & RTE & fps \\
\midrule
WiLoR~\cite{wilor} & DROID-SLAM & \xmark & 52.49 & 20.64 & 60.05 & 23.06 & 10.58 & 83.31 & 26.87 & 267.54 & 56.99 & 1.73 & 2.5 \\
HaWoR~\cite{hawor} & DROID-SLAM & \xmark & 31.96 & 13.79 & 36.45 & 14.91 & 7.52 & 76.44 & 28.55 & 210.46 & 53.62 & 2.22 & 4.9 \\
\rowcolor{fahrow} \mLV & DROID-SLAM & \xmark & 22.69 & 8.81 & 27.17 & 9.81 & 3.68 & 68.20 & 23.53 & 260.56 & 55.02 & 1.35 & 4.4 \\
\rowcolor{fahrow} \mLW & \paper-camera & \cmark & 26.60 & 10.87 & 32.24 & 12.62 & 4.73 & 55.90 & \best{21.26} & 177.65 & 45.76 & 1.30 & \best{38} \\
\rowcolor{fahrow} \mLW & DROID-SLAM & \xmark & \best{21.74} & \best{8.49} & \best{25.92} & \best{9.44} & \best{3.23} & \best{55.20} & 21.73 & \best{152.64} & \best{40.60} & \best{1.18} & 7.1 \\
\midrule
WiLoR~\cite{wilor} & GT & --- & 50.58 & 19.93 & 57.82 & 22.10 & 9.27 & 53.91 & 18.48 & 130.34 & 23.80 & 0.78 & --- \\
HaWoR~\cite{hawor} & GT & --- & 30.68 & 13.17 & 34.62 & 14.03 & 7.00 & 46.73 & 20.81 & 90.78 & 25.50 & 1.19 & --- \\
\rowcolor{fahrow} \mLV & GT & --- & \best{19.91} & \best{7.59} & \best{23.57} & \best{8.04} & \best{2.50} & \best{33.45} & \best{14.45} & \best{77.19} & \best{17.20} & \best{0.43} & --- \\
\bottomrule
\end{tabular}
\end{table*}

\begin{table}[t]
\centering
\caption{\textbf{Hand contact on held-out HOT3D and DexYCB s0 test.} Pooled vertex F1/AP and hand-level F1 ($\uparrow$), with 10\,mm contact labels. Ground-truth boxes unless marked $^\dagger$ (full pipeline); distance rules use known object geometry. HACO is zero-shot on HOT3D. Protocol and selection details: \cref{sec:benchmarks,sec:supp_chead}.}
\label{tab:mcontact}
\tabfont
\setlength{\tabcolsep}{3pt}
\begin{tabular}{lrrrrrr}
\toprule
 & \multicolumn{3}{c}{HOT3D} & \multicolumn{3}{c}{DexYCB s0} \\
\cmidrule(lr){2-4} \cmidrule(lr){5-7}
method & F1 & AP & flag & F1 & AP & flag \\
\midrule
HACO~\cite{haco} & 0.58 & 0.47 & 0.82 & 0.66 & 0.61 & 0.86 \\
\rowcolor{fahrow} \mS & 0.77 & 0.83 & \best{0.98} & 0.75 & 0.81 & \best{0.98} \\
\rowcolor{fahrow} \mBrel & 0.79 & 0.85 & \best{0.98} & 0.76 & \best{0.82} & \best{0.98} \\
\rowcolor{fahrow} \mL & \best{0.80} & 0.85 & \best{0.98} & 0.76 & \best{0.82} & \best{0.98} \\
\rowcolor{fahrow} \mH & \best{0.80} & \best{0.86} & \best{0.98} & \best{0.77} & \best{0.82} & \best{0.98} \\
\midrule
\rowcolor{fahrow} \mL, per frame$^\dagger$ & \best{0.80} & 0.85 & \best{0.98} & 0.76 & \best{0.81} & \best{0.98} \\
\rowcolor{fahrow} \mLV$^\dagger$ & \best{0.80} & \best{0.86} & 0.97 & \best{0.77} & \best{0.81} & 0.97 \\
\midrule
rule: dense points & 0.72 & --- & 0.98 & 0.56 & --- & 0.95 \\
rule: fitted MANO & 0.65 & --- & 0.96 & 0.56 & --- & 0.94 \\
\bottomrule
\end{tabular}
\end{table}

\begin{figure*}[t]
\centering
\includegraphics[width=\linewidth]{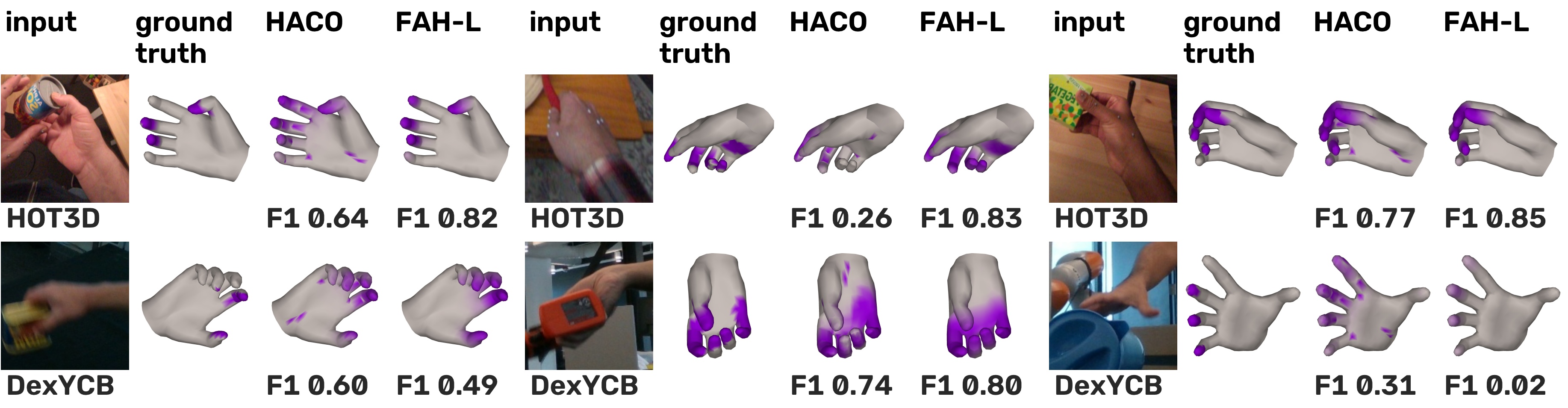}
\caption{\textbf{Contact on held-out HOT3D and DexYCB hands.} Each example shows the input, ground truth, HACO~\cite{haco} and \mL on the same ground-truth MANO mesh. Gray to violet indicates contact probability, with each method's validation threshold mapped to 0.5; scores are per-hand vertex F1. HACO wins on the fourth example; the last shows a missed fingertip contact. Selection: \cref{sec:supp_contact}; credits: \cref{sec:supp_credits}.}
\label{fig:mcontact}
\end{figure*}

\subsection{Contact}
\label{sec:m_cexp}
Without object geometry at inference, \mL outperforms HACO on HOT3D and DexYCB (\cref{tab:mcontact}).
The advantage also holds under HACO's per-hand averaging protocol: 0.76 vs 0.60 on HOT3D and 0.71 vs 0.66 on DexYCB.
\Cref{fig:mcontact} includes six examples, one favoring HACO and one FAH failure.

Accuracy grows from \mS to \mH on HOT3D, DexYCB and ARCTIC; on in-domain HOI4D, \mH is not above \mL (\cref{tab:supp_cfamily}).

On MOW~\cite{mow} under HACO's protocol and labels (35\,mm, per image), with both trained on MOW, \mS and \mBrel improve on HACO (F1 0.58 and 0.57 vs 0.52), while neither \mL nor \mH, the weakest of the four (0.53), is significantly above it.
On the pipeline's own detections, \mLV mainly improves contact timing: HOT3D onset and release vertex F1 rise by 0.053 and 0.055.
False contact on hovering HOT3D hands falls from 1.50 to 0.82\% of vertices, and DexYCB vertex F1 rises by 0.013.

% These gains have costs: hand-flag F1 falls by 0.004--0.008 on every set, and vertex F1 falls by 0.014 on ARCTIC and 0.006 on HOI4D (the latter interval includes zero).

\subsection{Hands in world space}
\label{sec:m_world}
\oursW places the video hands in world space with \paper-camera, our strictly causal camera model (\cref{sec:m_cam,tab:mworld}).
Errors are lower than HaWoR's offline pipeline on all reported metrics.
The 95\% sequence-bootstrap intervals exclude zero on HOT3D, EgoStandard chunks and RTE, but include zero for EgoStandard whole tracks; on frames detected by both methods, the whole-track intervals also exclude zero (\cref{tab:supp_tracks}).
To separate hand and camera effects, we also use HaWoR's DROID-SLAM camera with our hands.
This improves on HaWoR's hands except on EgoStandard whole tracks (W 260.56 vs 210.46\,mm; interval includes zero).
With our hands fixed, \paper-camera has higher HOT3D chunk W error than HaWoR's DROID-SLAM camera; its EgoStandard difference is unresolved.
The offline mode (DROID-SLAM with our hand masks and \paper-camera's scale) lowers EgoStandard error relative to HaWoR's DROID-SLAM camera, with no statistically resolved difference on HOT3D (\cref{sec:m_cam}).

\begin{table}[t]
\centering
\caption{\textbf{Dense points versus MANO parameters in one network} on HOT3D (3D joints 4.41\,mm PA-MPJPE): mesh fitted to the dense points (dense) vs the network's own MANO-parameter head (param.). 2D: median per hand vs the direct 2D joints, \% of hand size. Last rows: mesh vs the dense prediction, rigidly aligned (a diagnostic of the fit, not of accuracy).}
\label{tab:mdense}
\tabfont
\setlength{\tabcolsep}{4pt}
\begin{tabular}{lrr}
\toprule
 & dense & param. \\
\midrule
3D joints vs 2D (\%) & \best{1.43} & 3.07 \\
mesh joints vs 2D (\%) & \best{2.28} & 3.07 \\
mesh PA-MPVPE (mm) & \best{4.09} & 6.79 \\
mesh vs dense joints (mm) & \best{2.52} & 4.15 \\
mesh vs dense surface (mm) & \best{1.64} & 3.51 \\
\bottomrule
\end{tabular}
\end{table}

\begin{table}[t]
\centering
\caption{\textbf{Backbone and input resolution.} Controlled 12k-step ablation with one recipe and no external distillation: HOT3D PA-MPJPE (mm) / bench2d v2 PCK@0.1 / desktop fps, without the fingertip step.}
\label{tab:mres}
\tabfont
\setlength{\tabcolsep}{2pt}
\begin{tabular}{lccc}
\toprule
 & 256\,px & 320\,px & 448\,px \\
\midrule
ViT-S & 5.50/0.80/\best{117} & 5.40/0.80/109 & 5.35/0.80/91 \\
ViT-B & 4.82/0.82/104 & 4.75/0.82/93 & 4.70/0.82/73 \\
ViT-L & 4.53/\best{0.83}/80 & 4.46/\best{0.83}/68 & 4.41/\best{0.83}/47 \\
ViT-H+ & --- & --- & \best{4.34}/\best{0.83}/28 \\
\bottomrule
\end{tabular}
\end{table}

\subsection{Ablations and analysis}
\label{sec:m_analysis}
\label{sec:m_abl}
\paragraph{Dense points and mesh fitting (\cref{tab:mdense}).}
We compare a mesh fitted to dense points with a MANO-parameter head trained in the same network.
The parameter head disagrees more with the image (its joints lie 3.07\% of hand size from the direct 2D prediction, against 1.43\% for the dense 3D joints and 2.28\% for the fitted mesh's joints) and is less accurate (PA-MPVPE 6.79 vs 4.09\,mm).
Against SAM~3D Body's MHR decoder with a direct keypoint head, the dense head is better on HOT3D joints (4.38 vs 4.85\,mm), 2D (0.84 vs 0.83 PCK@0.1), root placement (8.67 vs 11.59\,mm) and silhouette agreement, at 1.7$\times$ the speed (\cref{sec:exp_dense}).
\paragraph{Backbone and resolution (\cref{tab:mres}).}
Larger encoders and inputs improve HOT3D accuracy in the controlled ablation; input resolution changes in-the-wild 2D PCK by at most 0.004. The released \mL uses ViT-L at 320\,px; on the desktop, speed falls from 117\,fps (ViT-S, 256\,px) to 28\,fps (ViT-H+, 448\,px).
\begin{table}[!htb]
\centering
\caption{\textbf{Part-supervision ablation.} Matched 2k-step joint training: PA-MPJPE (mm) on HOT3D and DexYCB for ViT-B; HOT3D PA-MPJPE and mined-set PCK@0.05 for continued \mL training. Union loss combines part-union/mask and dense-part terms; $\times$0.3 scales pose gradients.}
\label{tab:supp_parts}
\tabfont
\setlength{\tabcolsep}{2pt}
\begin{tabular}{lrr}
\toprule
ViT-B & HOT3D & DexYCB \\
\midrule
control & \best{4.43} & 5.69 \\
+ dense part terms & \best{4.43} & \best{5.68} \\
+ part head, union loss & 4.47 & 5.72 \\
\midrule
\mL, continued & HOT3D & mined \\
\midrule
without parts & \best{4.30} & 0.80 \\
+ joint part head ($\times$0.3) & 4.61 & \best{0.81} \\
\bottomrule
\end{tabular}
\end{table}

\begin{table}[t]
\centering
\caption{\textbf{Direct versus projected FAH outputs.} Internal comparison using a ViT-B development model on matched HOT3D hands, with true intrinsics. PCK@0.05 ($\uparrow$) compares direct 2D, projected camera-frame joints and projected mesh joints; PA-MPJPE ($\downarrow$, mm) compares direct 3D with mesh-regressed MANO joints plus tips.}
\label{tab:magree}
\tabfont
\setlength{\tabcolsep}{3pt}
\begin{tabular}{lrrrrr}
\toprule
 & \multicolumn{3}{c}{2D PCK@0.05 $\uparrow$} & \multicolumn{2}{c}{PA-MPJPE $\downarrow$} \\
\cmidrule(lr){2-4} \cmidrule(lr){5-6}
method & 2D & 3D proj. & mesh proj. & 3D & mesh \\
\midrule
FAH, image & \best{0.83} & \best{0.79} & \best{0.77} & 4.41 & 5.14 \\
FAH, video & \best{0.83} & \best{0.79} & \best{0.77} & \best{4.24} & \best{5.04} \\
\bottomrule
\end{tabular}
\end{table}

\paragraph{2D versus projected 3D.}
Projecting camera-frame 3D transfers errors in geometry and placement to 2D: in our ViT-B comparison the direct 2D head beats projected 3D and mesh joints (PCK@0.05 0.83 / 0.79 / 0.77; \cref{tab:magree}) and the dense 3D joints beat the mesh's regressor joints (4.41 vs 5.14\,mm), so we keep separate outputs despite their 1.42--1.78\% disagreement.
Further image/video ablations and data-mix studies appear in \cref{tab:abl_image,tab:abl_video,tab:mix}.

\section{Limitations and Conclusion}
\label{sec:m_concl}
\paragraph{Limitations.}
\label{sec:m_evalscope}
\paper has the following limitations:
\begin{itemize}
    \item Camera estimation dominates world-space error. Better camera trajectories remain an important direction (\cref{sec:supp_camera}).
    \item Contact without object geometry remains ambiguous, and most contact vertices are hidden on 96\% of HOT3D contact hands; the head also fires on hand-face and hand-hand contact in our in-the-wild videos, although training labels cover only hand-object contact. 
    \item \paper prioritizes real-time inference over maximum accuracy. Stronger models could improve its supervision and camera estimates: SAM~3~\cite{sam3} or vision-language-guided segmentation could replace SAM~2.1 pseudo-masks, and a stronger optical-flow model could replace RAFT-small in the camera motion head. These changes require evaluating accuracy gains against additional computation.
\end{itemize}
\paragraph{Conclusion.}
We present \paper, a fast yet accurate hand estimation method. It predicts corresponding 2D and 3D hand points and fits MANO or MHR-Hand in a separate stage, improving mesh accuracy over a parameter head in the same network.
The predicted surface also supports parts and contact; causal refinement and camera estimation extend it to video and world space. \paper achieves state-of-the-art performance on different hand estimation tasks while maintaining real-time speed on a consumer GPU.

{
    \small
    \bibliographystyle{unsrtnat}
    \bibliography{main}
}

\clearpage
\setcounter{page}{1}
\renewcommand{\thepage}{S\arabic{page}} % Keep supplement page links distinct from the main text.
\maketitlesupplementary
\appendix
\renewcommand{\topfraction}{0.95}\renewcommand{\dbltopfraction}{0.95}\renewcommand{\textfraction}{0.05}\renewcommand{\floatpagefraction}{0.7}\renewcommand{\dblfloatpagefraction}{0.7}\setcounter{topnumber}{5}\setcounter{dbltopnumber}{5}\setcounter{totalnumber}{8} % Supplement: denser float placement.

% Keep supplementary pages compact without stretching inter-paragraph gaps.
\raggedbottom
\setlength{\textfloatsep}{10pt plus 2pt minus 2pt}
\setlength{\dbltextfloatsep}{10pt plus 2pt minus 2pt}
\setlength{\floatsep}{10pt plus 2pt minus 2pt}
\setlength{\dblfloatsep}{10pt plus 2pt minus 2pt}
\setlength{\intextsep}{8pt plus 2pt minus 2pt}
\makeatletter
\setlength{\@fptop}{0pt}
\setlength{\@fpsep}{12pt plus 2pt minus 2pt}
\setlength{\@fpbot}{0pt plus 1fil}
\setlength{\@dblfptop}{0pt}
\setlength{\@dblfpsep}{12pt plus 2pt minus 2pt}
\setlength{\@dblfpbot}{0pt plus 1fil}
\makeatother

\section{Models and training}
\label{sec:supp_path}
\label{sec:method}
This section describes how we train the released models. Development ablations are in \cref{tab:abl_image,tab:abl_video}.
\label{sec:models}
\mMHR is the 320\,px ViT-B student with SAM~3D Body's decoder, MHR head and direct keypoints (\cref{sec:kphead}).
Unless noted, development ablations use the decoder-free ViT-B at the last milestone with YOLOv8m-seg, verification, tall-box crops, one re-crop and MANO fitting; \oursW runs with mesh fitting off (joints unchanged).

\subsection{Training setup}
\label{sec:supp_common}
We train pose networks with AdamW and weight decay 0.01, using a separate learning rate for the encoder. The schedule uses linear warm-up and cosine decay. We clip gradients at 1.0 and skip updates whose pre-clip gradient norm exceeds a threshold. Training sources are sampled with fixed weights independent of dataset size (\cref{tab:supp_data}).
We sample square windows from a 640\,px crop with context factor 1.25. Scale is sampled as $\exp(\mathcal{U}(\log 0.85,\log 1.15)\cdot j)$ and shift as $\mathcal{U}(\pm0.08)\cdot j$, with jitter $j=1$ unless noted. We do not rotate these crops.

Photometric augmentation changes contrast, brightness, saturation and white balance with probability 0.8. Their ranges are 0.6--1.4, $\pm30$, 0.5--1.5 and 0.9--1.1, respectively. Grayscale conversion, blur and JPEG compression have probabilities 0.05, 0.15 and 0.15; JPEG quality ranges from 30 to 80.

The student also receives low-light and egocentric degradations, while the teacher sees a clean 384\,px view. Low-light synthesis reduces exposure to 0.015--0.25 of the original and adds shot and read noise. Egocentric augmentation adds motion blur, sensor noise and 1--3 occluders, and reduces resolution by a factor of two to four. Hand-specific occluders cover the wrist or fingertips.

\begin{table*}[!tp]
\centering
\caption{\textbf{Pose training data} at the last development milestone: crops, labels and sampling shares. 2D/3D: joints; mesh: MANO vertices; mask: SAM~2.1; KD: teacher targets. Released stages add HaGRID pseudo-labels (4.9\%) and 885 hard crops (2.9\%; \cref{sec:supp_engine}). $^\dagger$Converted to the MANO convention (\cref{sec:labelconv}); $^\ast$benchmark sequences excluded.}
\label{tab:supp_data}
\tabfont
\setlength{\tabcolsep}{4pt}
\begin{tabular}{lrlr}
\toprule
source & crops & labels & share (\%) \\
\midrule
HOT3D train, every 2nd frame$^\ast$~\cite{hot3d} & 150k & 2D, 3D, mesh & 13.8 \\
COCO-WholeBody~\cite{cocowholebody}$^\dagger$ & 58k & 2D, mask, KD & 10.6 \\
CMU Panoptic hands~\cite{cmuhands}$^\dagger$ & 1.9k & 2D, KD & 3.2 \\
HInt train~\cite{hamer}$^\dagger$ & 12.4k & 2D, KD & 6.3 \\
Halpe-FullBody hands~\cite{halpe}$^\dagger$ & 21.7k & 2D, KD & 8.5 \\
FreiHAND~\cite{freihand}$^\dagger$ & 130k & 2D, 3D, mesh & 5.3 \\
EgoDex test split~\cite{egodex}$^\dagger$ & 148k & 2D, 3D (MANO convention) & 6.3 \\
EgoDex training part 1, filtered~\cite{egodex}$^\dagger$ & 565k & 2D, 3D (MANO convention), mask & 6.3 \\
VISOR~\cite{visor} & 17k & masks & 8.5 \\
background negatives & 84k & no hand & 5.3 \\
DexYCB train~\cite{dexycb} & 48.7k & 2D, 3D, mesh & 5.3 \\
InterHand2.6M / OakInk2 / HO-3D / HanCo~\cite{interhand,oakink2,ho3d,hanco} & 334k & 2D, 3D, mesh & 5.3 / 4.2 / 3.2 / 2.1 \\
OakInk2 egocentric, contact-first & 138.9k & 2D, 3D, mesh & 5.3 \\
hard examples (data engine, \cref{sec:supp_engine}) & 873 & 2D & 0.5 \\
\bottomrule
\end{tabular}
\end{table*}

\subsection{Teacher, students and the MHR keypoint model}
\label{sec:pose}
\label{sec:handonly}
\label{sec:finetune}
\label{sec:student}
\label{sec:kphead}
\paragraph{Hand-only SAM~3D Body teacher.}
We extract SAM~3D Body's~\cite{sam3dbody} hand branch and run it independently with bit-exact outputs. It contains a DINOv3-H+ encoder~\cite{dinov3}, promptable decoder, hand-mode MHR head and camera head, retaining 899M of the original 953M parameters. A fast rig operates on the hand sub-skeleton and its 2,445 vertices. We map these vertices to MANO joints with a sparse convex regressor with 169 weights, fitted on 4,070 FreiHAND training images, and mirror left hands.

We fine-tune the teacher for 10k steps on 384\,px crops, with 500 warm-up steps. The decoder and heads use a learning rate of $2\cdot10^{-5}$; the last eight encoder blocks and final normalization layer use one quarter of that rate. All other encoder layers remain frozen.
We sample HOT3D and COCO-WholeBody with weight 1, Halpe with 0.8, HInt and the EgoDex test split with 0.6, FreiHAND with 0.5, and CMU with 0.3. The EgoDex 3D supervision weight is 0.3.
The training objective is
\begin{equation}
\mathcal L=\mathcal L_{\mathrm{2D}}+3\mathcal L_{\mathrm{3D}}
 +0.3\mathcal L_{\mathrm{root}}+0.5\mathcal L_{\mathrm{KD}}.
\end{equation}
We use smooth-L1 penalties with $\beta=0.02$ for crop-normalized 2D coordinates and $\beta=0.01$\,m for 3D coordinates. The distillation targets come from the frozen off-the-shelf model. Results are in \cref{tab:abl_image}.
\paragraph{Feature distillation.}
The 840M-parameter backbone accounts for most of the computation. We distill it into DINOv3 ViT-S/16, ViT-B/16 and ViT-L/16 students through a two-layer adapter to the teacher's 1280-dimensional feature space. The students match teacher tokens from 384\,px views, area-pooled to $16\times16$, using variance-normalized MSE plus cosine distance. Distillation initializes the students; it is not their final training objective.
We train for 20k steps with AdamW, weight decay 0.05 and a learning rate of $2\cdot10^{-4}$, multiplied by five for the adapter. The schedule uses 1k warm-up steps followed by cosine decay. We rotate crops by up to $25^\circ$ with probability 0.7 and disable flips. Checkpoints are selected using the frozen decoder on FreiHAND evaluation images.
The ViT-B student then trains for 12k steps at 256\,px and 8k at 320\,px with SAM~3D Body's decoder, mask and verification heads.
\paragraph{Direct keypoints (\mMHR).}
A 1.6M-parameter head on the decoder's final tokens predicts 21 joints and 64 MHR surface points (4$\times$ upsampling, stride-4 RGB stem, soft-argmax 2D and heatmap-pooled 3D). It improves 2D alignment, particularly at fingertips (\cref{tab:abl_image}). Fitting the rig to these 2D predictions worsens 3D accuracy, so \mMHR uses the head's 2D and the mesh's 3D (\cref{sec:exp_dense}).

\subsection{Dense pose network}
\label{sec:dense}
\label{sec:supp_dense}
\label{sec:posedata}
\label{sec:egodex}
The stride-4 heatmap head also carries a pooled token with a CLIFF-style crop condition~\cite{cliff}; bone points lie at 1/3 and 2/3 of each bone, and the first dense models predicted all 778 vertices.
Least-squares translation is less stable in root depth than the MHR camera head. We therefore predict root depth explicitly, using both teacher-camera targets and ground-truth supervision.
\paragraph{Supervision.}
The point losses follow \cref{eq:point_loss}; KD sources are listed in \cref{tab:supp_data}.
We supervise surface points with ground-truth MANO vertices where available. Otherwise, SAM~2.1 pseudo-masks~\cite{sam2} at $96\times96$ pull projected points inside the blurred mask, excluding points that the teacher mesh identifies as occluded. A one-sided Chamfer term encourages coverage of the mask boundary. Teacher-mesh, edge-length and bone-anchor priors receive weights of 0.3, 0.3 and 0.5, respectively.
We cut masks a quarter hand length behind the wrist. We discard a mask if fewer than half the keypoints fall inside it or its IoU with the projected mesh is below 0.2.
For the reprojection loss in \cref{eq:reprojection_loss}, unlabeled joints receive half weight.

\paragraph{Surface basis and EgoDex filter.}
We project the 778-point predictions onto the lowest 256 Laplacian eigenfunctions to remove spikes. This basis reconstructs ground-truth HOT3D meshes to 0.61\,mm. The point-count ablation is in \cref{tab:npoints}.
EgoDex's ARKit labels~\cite{egodex} are tracker outputs. We retain the 58\% of 975k first-part crops whose 2D disagreement with \mMHR is below 0.12 hand sizes.
\paragraph{Milestone recipe (\cref{tab:abl_image}).}
Each stage resumes the previous checkpoint and uses a learning rate of $10^{-4}$ unless noted.
\begin{enumerate}
\item We first train the dense-point and root heads on MHR features for 20k steps at $2\cdot10^{-4}$, using ten sources and a weight of 0.6 for filtered EgoDex. We then train root depth for another 4k steps.
\item We add mesh and mask supervision, sampling every second HOT3D frame at $2.5\times$ weight. We train with EgoDex SAM masks for 7.1k steps, then add FreiHAND meshes for 8k steps.
\item We add reprojection consistency with weight 1 for 8k steps. Another 8k steps add DexYCB and visibility supervision, each at weight 0.5, plus spectral smoothness beyond 256 modes and a teacher-surface prior, each at weight 1. To train verification, we present 15\% of crops under the incorrect handedness hypothesis.
\item For label harmonization, we sample InterHand2.6M, OakInk2, HO-3D and HanCo at weights 0.5, 0.4, 0.3 and 0.2, respectively, and hard examples at 0.03. Egocentric degradation and occluder probabilities are 0.25 and 0.2; box jitter is 1.6, and surface reprojection has weight 0.5. This stage runs for 8k steps from the intermediate checkpoint with higher EgoDex 3D weight.
\item We initialize the released heads with 400 points and 200 spectral modes and train for 8k steps. Sampling weights are 1.3 for HOT3D, 0.05 for hard examples, and 0.5 each for egocentric OakInk2 and EgoDex with WiLoR shape targets.
\end{enumerate}

\subsection{Label conventions}
\label{sec:labelconv}
EgoDex's ARKit supervision biases finger lengths and wrist position in egocentric images, including for the H+ teacher. Post-hoc bone and wrist corrections trade EgoStandard gains for HOT3D losses, so we correct the training labels.
We re-convert EgoDex against a kinematic WiLoR fit that never saw EgoDex; an EgoDex-trained reference would retain the ARKit proportions.
We use kinematic MANO joints throughout and refit the teacher's joint regressor. FreiHAND's mesh-regressed joints differ from this convention by 3.3\,mm on average and 6.9\,mm at the pinky MCP.
For 2D labels, we apply fixed corrections of $-10.8\%$ to Halpe knuckles, $-6.5\%$ to the COCO-WholeBody wrist and $+5.6\%$ to its thumb MCP.
On 2D-only data, WiLoR's MANO-native 3D provides the shape target, while the teacher's image-aligned mesh supplies the 2D surface targets. We also add 334k MANO-labeled crops from InterHand2.6M~\cite{interhand}, OakInk2~\cite{oakink2}, HO-3D~\cite{ho3d} and HanCo~\cite{hanco}, together with student-side egocentric augmentation and a surface reprojection loss.
Label conventions alone changed EgoStandard PA-MPJPE by 1.06\,mm.

\subsection{Part segmentation}
\label{sec:supp_parts}
The 17 classes follow MANO's skinning joints; the head adds $2{\times}2$ pooling and $3{\times}3$ convolutions to the design of \cref{sec:m_contact}.
We render part labels from ground-truth MANO meshes on HOT3D, DexYCB, egocentric OakInk2, HanCo and FreiHAND. We clip them to SAM~2.1 masks, retaining masks with silhouette IoU of at least 0.7. On COCO-WholeBody, CMU and HInt, we render our predicted mesh instead and ignore fingers that disagree with the 2D labels. These rendered labels agree better with HOT3D reference labels than nearest-bone assignment.
The head trains for 16k steps with \cref{eq:part_loss}. Joint-training ablations are in \cref{tab:supp_parts}.

Labels clipped to SAM~3~\cite{sam3} masks change part mIoU by at most 0.009 per dataset and lower boundary F against the SAM masks by 0.015--0.021, so the released heads keep the SAM~2.1 labels.

\paragraph{Parts, points and masks.}
On five held-out datasets, \mL achieves 0.46--0.54 mIoU over 16 parts; counting misses as background lowers COCO by 0.03--0.04 and the other sets by at most 0.003. The foreground union matches SAM masks at IoU 0.77--0.85.
Joint training slightly worsens pose accuracy (\cref{tab:supp_parts}), so we train the released part head on frozen pose features.

\paragraph{Dense-point finger assignment.}
\label{sec:supp_pointparts}
We compare matched ViT-B runs on ground-truth-mesh/SAM-labeled crops from HOT3D, DexYCB and OakInk2. Each run uses 300 crops per dataset and 2k training steps. Part losses reduce wrong-finger assignments from 6.7\% to 4.2--4.5\%, without a clear pose-accuracy gain. We stop gradients through validity weights and omit targets for absent parts; controls use one seed and corrected runs use two. Coarse finger-region losses improve label agreement further but cover fewer points, so they are not released.

\paragraph{Reference-label limitations.}
Review of 168 selected difficult HOT3D crops found 18 occluder-label errors and 3 misalignments; this sample cannot estimate population error. Agreement with these labels does not establish anatomical accuracy.

\subsection{Contact head}
\label{sec:supp_chead}
\paragraph{Architecture and training.}
The contact branch pools heatmap features by a factor of four and copies the encoder's final normalization layer. Winding numbers identify vertices inside the object for the contact labels.

We sample ARCTIC and HOI4D at weight 0.5 each and MOW at 0.05, holding out 16.5k crops for validation. HOI4D's Chair category is excluded because its ground truth is unreliable and often penetrates the object. All HOI4D left hands belong to this category, leaving only right hands. We also reject crops with hand penetration beyond 15\,mm on HOI4D or 25\,mm on other sources.

The contact head uses a learning rate of $3\cdot10^{-4}$.
\paragraph{Selection.}
For each size, we compare branch learning rates of $3\cdot10^{-5}$ and $10^{-4}$. We reject settings with a statistically significant vertex-F1 loss on any test set, then choose the best validation F1. Every size selects $10^{-4}$. Test scores therefore inform model selection.
\paragraph{Video model.}
In addition to the inputs in \cref{sec:m_contact}, the video model receives detection confidence, presence and side-normalized root position, with causal velocity and acceleration. It uses pre-norm blocks and four attention heads. Zero-initialized residuals initially reproduce the image model, and the hand flag resets after 32 absent frames.

We train for 20k AdamW steps at a learning rate of $5\cdot10^{-4}$, using warm-up and cosine decay. Batches contain 256 windows sampled at strides 1--3. Balanced point BCE and half-weight flag BCE form the loss. Detection dropout and noise on logits and motion inputs provide augmentation.

To obtain out-of-fold inputs, we train two image heads for 8k steps each. Each excludes half the HOT3D, DexYCB, HOI4D and ARCTIC recordings and predicts tracks for that held-out half. Inference uses the released image head. Separate validation recordings select the map blend and vertex recalibration. Flag hysteresis uses on/off thresholds of 0.90/0.10, except that \mS turns on at 0.55.

Training-source selection also uses test scores: we first require no significant vertex-F1 loss on HOT3D or DexYCB, then minimize loss on HOI4D and ARCTIC. This selects HOT3D and DexYCB for \mBrel, \mL and \mH, with HOI4D and ARCTIC added for \mS.

\subsection{Detectors}
\label{sec:supp_det}
\label{sec:detection}
Detection recall limits every downstream pose estimator. With WiLoR's detector, bench2d recall is 0.92 overall and 0.84 on COCO.
We train our detectors with Ultralytics~\cite{yolov8} and a cosine schedule. Horizontal flips are disabled to preserve handedness, and mosaic augmentation stops for the last three epochs.
\paragraph{Development pose detectors and verification.}
We fine-tune WiLoR's YOLOv8m-pose on hand annotations from COCO-WholeBody, VISOR, HOT3D, HInt and CMU. VISOR contributes hand-only boxes from frames where every hand is matched. Training uses 68.4k frames with 124.4k hands, 15 epochs at 640\,px and batches of 128. Inference uses 800\,px to retain small hands. Retraining substantially improves held-out detection accuracy (\cref{tab:det}). Adding synthetic low-light copies to YOLOv8s-pose preserves recall.

Verification combines detection confidence and hand presence by their geometric mean. Identity NMS then multiplies this score by the handedness score and ranks all hypotheses, including both sides of re-tried boxes. It suppresses a hypothesis if its wrist lies within half a palm length of an accepted wrist, where palm length runs from wrist to middle-finger MCP. This retains distinct hands in overlapping boxes without additional pose crops.

\paragraph{One-stage segmentation detectors.}
We initialize YOLOv8s-seg and YOLOv8m-seg from the pose detectors and train for 12 epochs at 800\,px on 90.2k frames. SAM~2.1-L supplies pseudo-masks, prompted by a box and up to eight visible keypoints. We clip masks to the box and replace failed masks with an inscribed ellipse.

We then fine-tune YOLOv8m-seg for six epochs using boxes that enclose both the mask and keypoint skeleton. Hard-example training images are oversampled fourfold. This improves finger coverage and pose accuracy, with a small loss in precision (\cref{tab:det}).

The released detector uses SAM~3~\cite{sam3} masks, with SAM~2.1 or ellipse fallbacks. Text prompts add 15.9k previously unlabeled hands. To correct merged overlapping hands, we split masks along the hand skeletons and recover missing second hands. We also add two-hand training frames from HaGRID (2,938) and HOT3D/OakInk2 (4,003). We exclude 15 HOT3D training sequences and all subjects used in our test sets.

These changes improve masks and overlapping-hand recall, with some loss on small, difficult hands. Segmentation results are in \cref{tab:masks}.

\subsection{Hard-example data engine}
\label{sec:supp_engine}
We mine Open Images and HaGRID for cases where models disagree. The candidate pool includes our model under both handedness hypotheses and multiple rotations, \mMHR, WiLoR, RTMPose-m and MediaPipe. Reviewers inspect candidate sheets under fixed rules and retain hands where our error exceeds 0.1 hand sizes. The review protocol includes repeat checks of retained hands or a random 10\% sample. Holdouts are separated by person on HaGRID and by image on Open Images.

The released training pool contains 885 crops with 2D labels and no teacher targets. It is sampled at weight 0.3, accounting for 2.9\% of training examples. The released hard\_mined\_v2 benchmark retains the original 169 held-out hands. This training improves the stress test without changing public-benchmark accuracy.
\paragraph{HaGRID pseudo-labels.}
We also use consensus 2D labels on 165,118 HaGRID crops covering 33 gestures. They account for 4.9\% of training samples and receive teacher targets at weight 0.5.

An earlier dense model, \mMHR's keypoint head and RTMPose-m provide candidates. A joint is labeled only when at least two agree within 0.05 hand sizes; the label is their median. We retain sharp, uncropped hands with at least 12 labeled joints, boxes of at least 100\,px, and handedness agreed by two models. People in the hard-example benchmark are excluded. Neither training pool is released.

\subsection{Mesh stage}
\label{sec:supp_mesh}
\label{sec:meshfit}
\paragraph{Closed-form initialization.}
We split the template into 16 parts by largest skinning weight. Horn's method~\cite{horn1987} estimates each part's rotation from the centered template and predicted points. Parent-relative rotations $R_\text{parent}^\top R_\text{child}$ give the initial 45-dimensional pose (\cref{tab:meshfit}).
\paragraph{Learned fitter.}
An MLP predicts pose, shape and root corrections from the initial pose, 195 surface points and 21 joints. Points and joints are wrist-centered in the root frame. The MLP has three 1024-wide GELU layers and a zero-initialized output.

We train it on the development model's predictions for HOT3D and DexYCB training clips, using L1 losses on vertices, joints and centered vertices. Training uses AdamW at $10^{-3}$, batches of 512 and 20 epochs over 124.9k hands. Input noise of 3\,mm is necessary for transfer (\cref{tab:meshfit}); this setting was fixed before testing.
\paragraph{Gauss--Newton step.}
One damped Gauss--Newton step refines the 61 parameters in fp32. It uses a Huber penalty with $\delta=1$\,cm, weighting the 778 vertices by 0.3 and the 21 joints by 1. We compute the central-difference Jacobian with step $10^{-2}$, batching all perturbed parameters in one compiled MANO pass. This matches the original mesh to within 0.004\,mm.

Initialization, the learned fitter and refinement account for 37\%, 42\% and 21\% of mesh-stage runtime, respectively. Rigid PnP aligns overlay meshes to the 2D joints.
\paragraph{MHR-Hand fit.}
The same dense points can drive an MHR-Hand fit using a learned initializer and one damped Gauss--Newton step with joint limits. Our hand-mode rig matches SAM~3D Body's TorchScript MHR to within 0.001\,mm. The joint regressor is fitted on training splits only, and the fit exports native SAM~3D Body/MHR parameters (\cref{tab:meshfit}). On HOT3D, \mL runs at 65\,fps with either fit or 59\,fps with both.

\subsection{Video refiner}
\label{sec:supp_refiner}
\label{sec:refiner}
\paragraph{Model and training.}
The causal pre-norm transformer has four attention heads. Each frame contributes a 110-dimensional vector containing presence, detection confidence, root-relative 3D joints, 2D joints and root position as $(x/z,y/z,\log z)$. Missing inputs are zero-filled.

Training uses 600 HOT3D and 1,600 DexYCB clips. The loss is $e_\text{rel}+e_\text{PA}+3e_\text{acc}+0.1e_\text{root}$, with errors in mm. Track augmentation includes mirroring, in-plane rotation, image shift, log-depth offset, detection dropout and noise. AdamW training runs for 20k steps at $5\cdot10^{-4}$ with batches of 256 windows. All released models share this refiner.
\paragraph{Inference.}
We fill gaps of at most two frames. The One Euro filter uses a 12\,Hz cutoff and $\beta=1$; handedness memory is 0.8. A separate filter on direct 2D predictions avoids errors from projecting refined depths. It uses 30\,Hz and $\beta=0.05$ per px/s, selected on HOT3D and DexYCB. The mesh follows the refined joints.
\paragraph{Track proposals.}
After ten confirmed detections, a track can propose a constant-velocity box. Presence and pose continuity must support the proposal, and detector boxes always take priority. We disable blind gap filling when no hand is found or less than 60\% of the box lies inside the image. Track proposals reduce missed hands and false positives on EgoStandard with little change in pose accuracy or throughput.

\subsection{Camera model}
\label{sec:supp_camera}
\label{sec:camera}
World-space error combines hand error with camera rotation, trajectory and metric-scale errors, whose contributions depend on the dataset and camera mode (\cref{tab:supp_camdiag,sec:exp_world}).
The reference is HaWoR's offline pipeline~\cite{hawor}: masked DROID-SLAM~\cite{droidslam} at 440\,px on every frame with Metric3D~\cite{metric3dv2} scale on every keyframe.
\paragraph{Causal mode.}
We estimate each arriving frame's camera pose by combining DPVO, metric depth and a learned motion head.

DPVO~\cite{dpvo} runs at 384\,px on alternate frames, masking the previous hand boxes. We reinitialize from eight buffered frames after a rotation step above $14.2^\circ$ or a twofold scale jump. The rotation threshold is the 99.9th percentile on HOT3D training clips. Translation is disabled for pure rotations.

MoGe-2~\cite{moge2} estimates depth in a background process every 30 processed frames, or every 15 frames per DPVO instance. Each call takes about 0.1\,s and is applied five frames later. We correct depth scale by 1.034, fitted on 60 HOT3D training clips, and aggregate estimates with a running median in log space to reduce noise.

The motion head is a six-layer transformer. It combines two frames' DINOv3 ViT-S/16 tokens at 192$\times$256 resolution with pooled RAFT-small~\cite{raft} flow, intrinsics, frame-slot embeddings and hand-box embeddings. Outputs are rotation, metric translation and their log-variances. Training uses HOT3D, EgoDex part~2, egocentric ARCTIC~\cite{arctic} and HOI4D~\cite{hoi4d}, excluding subject H3. We sample strides 1--6 and lags up to 80 frames, with Laplace likelihood and a chained-prediction loss.

Fusion normally retains DPVO's rotation and corrects its translation scale using depth and the motion head's 30-step path length. The path-length correction weakens as DPVO's translation evidence increases. We measure this evidence as the median difference between rotation-only and full-pose reprojection residuals over 30 steps. The correction exponent falls from 0.5 to zero as evidence rises from 0.5 to 1.5\,px.

We use the motion head's step if DPVO loses tracking or their rotations differ by more than three times the predicted rotation error. We can also replace translation alone while keeping DPVO's rotation. This happens when translations differ by more than ten times the head's predicted error. The threshold drops to three times when hand boxes leave less than 70\% of the patch grid unmasked, because the remaining patches often lie on hand edges or moving objects.

\paragraph{Offline mode.}
Masked DROID-SLAM uses our hand masks to estimate the trajectory. We take the geometric mean of its Metric3D scale and the causal scale estimate. If their Sim3 disagreement exceeds 5\% of path length, we fall back to the causal estimate.
We report three runs, pairing SLAM realizations with causal seeds. SLAM scale varies by 10--20\% between runs. The full offline pipeline runs at 7.1\,fps (\cref{tab:mworld}).

\paragraph{Selection and test exposure.}
We develop camera settings on disjoint EgoStandard development clips and HOT3D training clips. We also check the 98 HOT3D benchmark clips and EgoDex part~2. Offline development uses non-benchmark HOT3D clips.

The pre-fixed acceptance rule requires at least a 2\% mean whole-track W and WA gain across domains, no worsening on chunks, and no domain loss above two seed standard deviations. The motion head was adopted for its development-set WA gain despite failing the W criterion.

HOT3D benchmark scores informed camera choices. Earlier configurations were also evaluated on EgoStandard test before the development split existed, so these are not untouched-test estimates. Later configurations were committed before testing. In particular, the translation rule was selected on development data and evaluated once on test.

Near-static cameras remain difficult: odometry can drift and maps can collapse. A few such tracks can reverse camera rankings between development and test (\cref{tab:supp_worldsd}).

\subsection{Inference pipeline and timing}
\label{sec:supp_systems}
\label{sec:speed}
\label{sec:supp_timing}
The eager decoder is limited by kernel launches and host-device synchronization. We reduce deployment overhead through TensorRT inference, shared frame uploads, GPU crop transforms and packed copies (\cref{tab:abl_image}). PyTorch remains available as a fallback.

Detection uses confidence 0.05, a fused threshold of 0.25 and wrist-based identity NMS. We train at 800\,px and infer at 640\,px in FP16. Pose crops span 1.1 times the box's long side, with one re-crop when a hand reaches the border.

\paragraph{Timing.}
Image and video timing uses the desktop (i7-14700F, WSL2, 8 threads) on 2,400 pre-decoded frames: the first 300 of four EgoStandard clips (1920$\times$1456) and four HOT3D streams (1408$\times$1408).
Loading, decoding and 48 warm-up frames per resolution are excluded; uninstrumented runs synchronize once per frame, and separate per-stage synchronization gives the shares.
fps is 1000 divided by the slower of two median frame times (ms); dataset-specific tables use their own frames, and no older-GPU timing is substituted.
World timing includes the camera stages on four EgoStandard clips, excludes loading, decoding and 30 warm-up frames, and uses the slower swap-free round.
Detector and encoder ONNX exports use mixed FP16 with fp32 I/O, built per GPU for TensorRT~11.4; the ViT-L engine is 637\,MB.

% The method details are merged into sec/s0_supp_head.tex (Appendix A); the earlier text is in the git history.

\section{Benchmarks and baselines}
\label{sec:benchmarks}
Crop-based baselines use their official fp32 implementations; our models use the optimized runtime (timing: \cref{sec:supp_timing}).

\paragraph{HOT3D v1 (3D).}
We evaluate on 98 five-second clips from the 27 validation sequences in HaWoR's split of HOT3D train\_aria~\cite{hot3d}. These clips are disjoint from our training data and contain 14.7k frames with 25.4k MANO hands. Aria RGB images are rectified to an upright pinhole camera.

We match detections at box IoU $\geq0.3$. Pose, mesh, root, relative-hand-position and acceleration errors are computed on matched hands; recall covers all ground-truth hands. Tables abbreviate PA-MPJPE as PA and PA-MPVPE as PVE. We normalize 2D acceleration error and jitter by hand size and compare consecutive matched frames with projected ground truth.

Following ViDiHand~\cite{wang2026vidihand}, PA-MPJPE-p (PA-p) penalizes missed hands. Each miss receives the unaligned error of a canonical, wrist-anchored, mean-shape MANO hand, approximately 130\,mm.

\paragraph{Hands in world space.}
We transform camera-frame joints into world space using predicted camera poses. For each ground-truth hand, we split its matched detections into contiguous runs and discard runs shorter than 16 frames. Each method is evaluated on its own matched frames, so detection misses and run filtering can change the scored population.

For chunk metrics, we divide runs into 100-frame windows and retain the remainder. W-MPJPE100 uses a similarity alignment fitted on the first two frames; WA-MPJPE100 fits the whole window. RTE uses a rigid alignment of the wrist path and normalizes its error by ground-truth path length. All three metrics are frame-averaged. Whole-track metrics use the same definitions over each complete run.

The shared reference camera comes from HaWoR's masked DROID-SLAM~\cite{droidslam} pipeline with Metric3D~\cite{metric3dv2} scale. W-MPJPE grows with alignment length, whereas path-normalized RTE tends to decrease with run length. We therefore compare RTE only within a dataset.

\paragraph{bench2d v2 (2D).}
This benchmark contains 6.4k images and 8.6k hands from HOT3D, CMU Panoptic, COCO-WholeBody validation and the HInt test splits. We report matched-hand PCK@0.1, fingertip PCK, recall and PCK including misses. Precision is scored only on fully annotated HOT3D; COCO annotations are not exhaustive.

The \emph{3D fit} metric measures projected 3D compatibility with visible 2D landmarks. We fit translation to the visible ground-truth joints, then compute PCK@0.1 using dataset intrinsics or an image-diagonal focal length. This is not a measure of metric depth or 3D accuracy. Crop-based mesh baselines use their own crop-space 2D predictions, which are more accurate than their reprojected meshes.

\paragraph{Custom sets.}
Joint and fingertip conventions differ across public datasets (\cref{sec:labelconv}). We use a common model-assisted review procedure for the challenging-pose, mined, low-light and shadow sets. Reviewers inspect all methods' candidates on original and brightened images. If none is usable, they prompt our model with a box under both handedness hypotheses. Doubtful cases receive further review, and decisions are stored with the labels.

These sets informed development and may favor methods in the candidate pool. We report recall, PCK@0.05 and PCK@0.1, fingertip PCK@0.1, PCK including misses and handedness accuracy. Matching requires mean visible-joint distance within half a hand size.

\begin{figure*}[t]
  \centering
  % The 72-dpi source is 1632 x 816 pixels: display its two 408-pixel rows side by side.
  \includegraphics[width=0.498\linewidth,trim={0 408bp 0 0},clip]{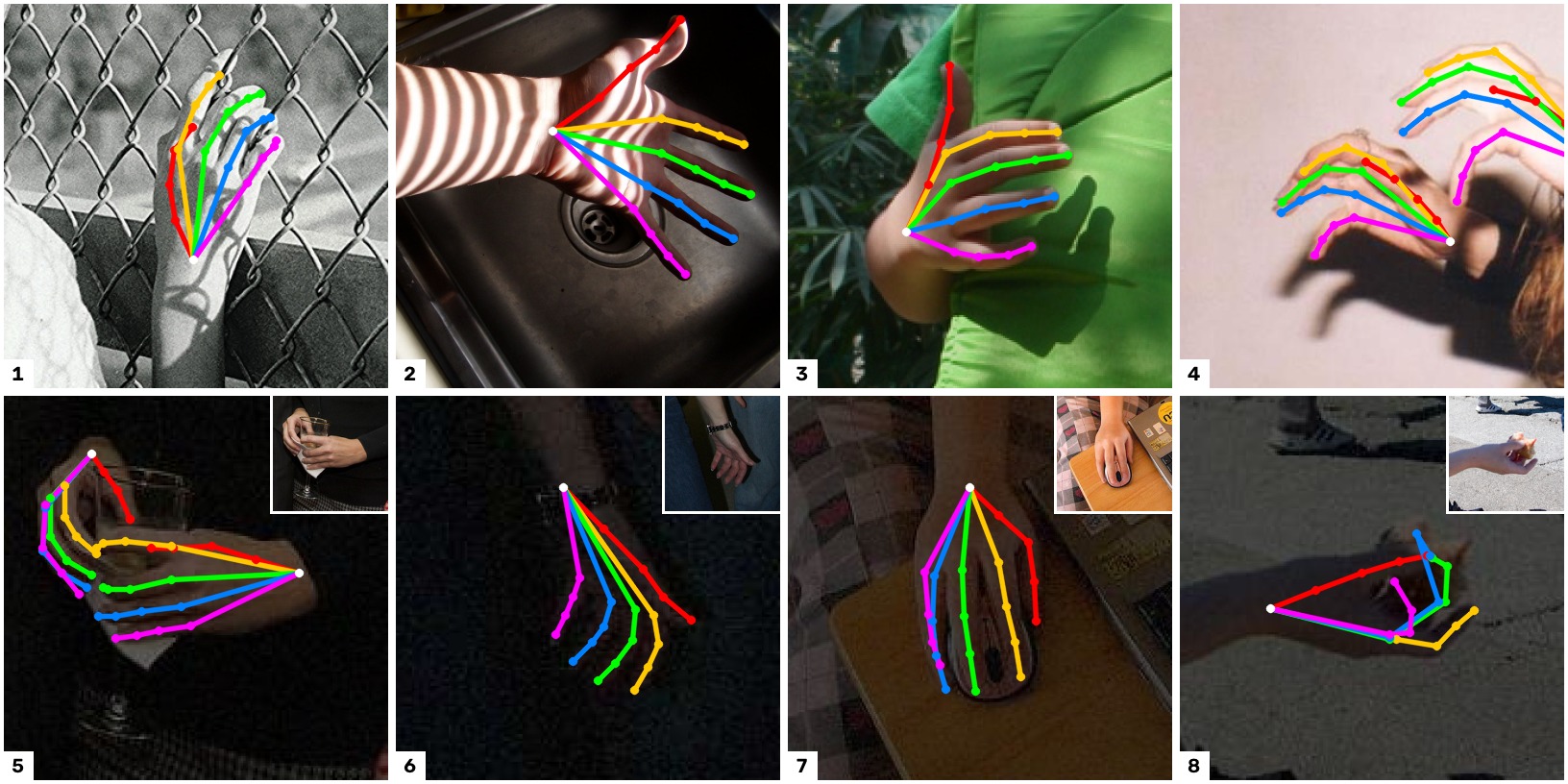}\hfill
  \includegraphics[width=0.498\linewidth,trim={0 0 0 408bp},clip]{figures/supp_bench_samples_v1.jpg}
  \caption{\textbf{Custom benchmark examples.} Images 1--4 show shadows across the hand or beside it; 5--8 show synthetic low light with the original image inset. Ground-truth keypoints are overlaid. Challenging-pose stock photos are omitted because their licenses do not permit redistribution. All shown photos are CC~BY~2.0; sources, authors and modifications: \cref{sec:supp_credits}.}
  \label{fig:supp_samples}
\end{figure*}
\paragraph{Challenging poses (hard\_pose\_v1).}
The challenging-pose set contains 175 hands on six stock-photo sheets covering tool grips, gestures, fists and sign-language poses. Labels start from a five-method consensus. We re-label contested or rarely detected hands using box prompts and review every label.
\paragraph{Hard in-the-wild examples (hard\_mined\_v2).}
This set contains 169 hands from held-out data-engine images (\cref{sec:supp_engine}) where an earlier model erred by more than 0.1 hand sizes. Most failures involve truncated fingers or incorrect handedness. It is a failure-mined stress test.
\paragraph{Low light (dark\_v1).}
We evaluate on 19 real dark photos with 50 hands (one unlabeled), plus a synthetic low-light version of bench2d v2. Synthesis preserves the original labels with exposure reduced to 0.05--0.15 of the original in linear light. We also add sensor noise, color shifts, gamma changes, blur and JPEG compression (\cref{fig:supp_samples}). The released set contains only the synthetic images because the real photos have unknown licenses.
\paragraph{Shadows (shadow\_v1).}
We collect images with shadows across the hand or beside it from Open Images, HaGRID~\cite{hagrid}, Openverse and Wikimedia Commons. CLIP~\cite{clip} ranking and manual review yield 170 images with 278 scored hands. Labels are per-joint medians of reviewer-selected predictions from our model, \mMHR, WiLoR, HaMeR and RTMPose.

\paragraph{EgoStandard.}
EgoStandard~\cite{egostandard} provides head-camera manipulation videos with hardware-assisted tracking of both hands and camera poses. The test set takes one 30\,s clip per test task (tasks chosen by a hash of the task name) from the left head camera: 120 clips, 105.9k frames.
Targets are MANO fits to the tracked keypoints, boxes are projected meshes, and world-space metrics use each method's own true positives; no frames are shown.
A disjoint dev split (80 clips from 182 other tasks) selects camera-model and track-proposal settings.
Earlier pose milestones, label-convention studies and camera configurations were scored on the test set, so it is not an untouched test set (\cref{sec:supp_camera}); each final released pose model was scored once, without selection among the four models on test scores.

\paragraph{DexYCB (third person).}
We use the seen-subjects (s0) test split of DexYCB~\cite{dexycb}. Each sequence is evaluated once, with cameras chosen to balance camera and object frequencies. The resulting set has 160 clips and 9.3k scored hands. We detect and match hands as on HOT3D rather than using ground-truth boxes.

\paragraph{Contact.}

\Cref{tab:mcontact} evaluates contact with ground-truth boxes. It uses 5,172 HOT3D hands sampled every fifth frame and 9,506 DexYCB s0 test hands. HACO is zero-shot on HOT3D and trained on DexYCB s0 train.

Full-pipeline evaluation includes 25,431 HOT3D and 9,319 DexYCB hand-frames. Object-distance rules use the detected subsets, with 25,378 and 9,225 hands, respectively. Each FAH size has one contact model with validation-calibrated thresholds. These evaluation sets informed model selection (\cref{sec:supp_chead,sec:supp_contact}).

Ground-truth contact marks MANO vertices within 10\,mm of or inside the ground-truth object mesh (winding number; any penetration depth); thresholds use separate validation data (\cref{sec:supp_chead}).
Pooled vertex F1 aggregates all hands, HACO's per-hand vertex F1 averages over contact-positive hands, and hand F1 scores the separate flag.
Onset and release windows lie within three frames of a switch, with five stable frames on each side; hovering means no contact at 10--30\,mm object distance; flips count flag switches.

\paragraph{EgoDex part~2 (model selection) and VISOR.}
We use 4,000 crops from EgoDex's second training part~\cite{egodex} for pose-model selection. No pose network trains on these crops. We convert labels to kinematic MANO joints and score PA-MPJPE with ground-truth boxes. VISOR~\cite{visor} validation frames with hand-plus-arm masks evaluate segmentation.

\paragraph{Baselines.}
WiLoR~\cite{wilor}, HaMeR~\cite{hamer}, Hamba~\cite{hamba} (on its initial-regressor joints) and the AnyHand~\cite{anyhand} checkpoints of WiLoR and HaMeR (released configurations) run per frame with WiLoR's detector (confidence 0.3, NMS 0.5) and the true intrinsics; HaWoR~\cite{hawor} runs on whole clips.
MediaPipe HandLandmarker~\cite{mediapipehands} (MediaPipe Hands in the tables) runs in image mode with up to four hands; its metric hand-centered landmarks are rotated to the viewing ray and placed by least squares onto its 2D.
The RTMPose-m hand model~\cite{rtmpose} (RTMDet-nano at 320\,px, ONNX Runtime) and the whole-body RTMW-l~\cite{rtmw} (YOLOX-m person detector) are 2D only.
EgoHandICL~\cite{xie2026handicl} reports P-MPJPE 4.0\,mm on its own ARCTIC split, on detected hands only, with no released model or test lists, so it is not in our tables.

\section{Extended results}
\label{sec:experiments}
\label{sec:supp_tables}

This section expands the main results with development ablations and design studies. A dash indicates an unavailable result. Unless noted, throughput is measured on HOT3D with an RTX 4070 Ti SUPER, using the slower of two rounds (\cref{sec:supp_timing}). For SAM~3D Body, $^\S$marks our CUDA-graph implementation; eager inference runs at 5.9\,fps.

\subsection{Image model}
\label{sec:exp_image}

\begin{table}[!htbp]
\centering
\caption{\textbf{Additional HOT3D metrics.} Detection recall and root-relative MPJPE complement the primary errors in \cref{tab:m3d}.}
\label{tab:hot3d}
\tabfont
\setlength{\tabcolsep}{4pt}
\begin{tabular}{lrr}
\toprule
method & recall $\uparrow$ & rr (mm) $\downarrow$ \\
\midrule
\multicolumn{3}{l}{\emph{per frame}} \\
MediaPipe Hands~\cite{mediapipehands} & 0.58 & 54.43 \\
HaMeR~\cite{hamer} & 0.96 & 57.03 \\
HaMeR + AnyHand~\cite{anyhand} & 0.96 & 49.56 \\
Hamba~\cite{hamba} & 0.96 & 56.49 \\
WiLoR~\cite{wilor} & 0.96 & 17.80 \\
WiLoR + AnyHand~\cite{anyhand} & 0.96 & 19.81 \\
SAM~3D Body hand~\cite{sam3dbody} & 0.96 & 28.30 \\
\rowcolor{fahrow} \oursI & \best{1.00} & \best{7.20} \\
\midrule
\multicolumn{3}{l}{\emph{video}} \\
MediaPipe Hands (video mode) & 0.82 & --- \\
HaWoR~\cite{hawor} & 0.93 & 14.21 \\
\rowcolor{fahrow} \mLV & \best{1.00} & 7.38 \\
\bottomrule
\end{tabular}
\end{table}

\begin{table*}[!tp]
\centering
\caption{\textbf{Additional bench2d v2 metrics.} We report fingertip and per-dataset PCK@0.1, overall recall and recall on overlapping hands. Aggregate PCK and accuracy including misses are in \cref{tab:m2d}.}
\label{tab:bench2d}
\tabfont
\setlength{\tabcolsep}{2.5pt}
\begin{tabular}{lrrrrrrrr}
\toprule
method & tips & HOT3D & CMU & COCO & HInt-ND & HInt-EPIC & recall & overlap recall \\
\midrule
MediaPipe Hands~\cite{mediapipehands} & 0.59 & 0.61 & 0.73 & 0.64 & 0.66 & 0.68 & 0.51 & 0.43 \\
RTMPose-m~\cite{rtmpose} & 0.67 & 0.69 & 0.73 & \best{0.75} & 0.72 & 0.72 & 0.56 & 0.31 \\
RTMW-l~\cite{rtmw} & 0.60 & 0.64 & 0.66 & 0.69 & 0.71 & 0.72 & 0.94 & \best{0.94} \\
HaMeR~\cite{hamer} & 0.64 & 0.72 & 0.73 & \best{0.75} & 0.72 & 0.66 & 0.92 & 0.80 \\
HaMeR + AnyHand~\cite{anyhand} & 0.60 & 0.72 & 0.67 & 0.71 & 0.64 & 0.58 & 0.92 & 0.80 \\
Hamba~\cite{hamba} & 0.64 & 0.72 & 0.72 & \best{0.75} & 0.72 & 0.69 & 0.92 & 0.80 \\
WiLoR~\cite{wilor} & 0.64 & 0.65 & 0.72 & \best{0.75} & 0.70 & 0.69 & 0.92 & 0.80 \\
WiLoR + AnyHand~\cite{anyhand} & 0.65 & 0.66 & 0.72 & \best{0.75} & 0.70 & 0.70 & 0.92 & 0.80 \\
SAM~3D Body hand~\cite{sam3dbody} & 0.61 & 0.69 & 0.53 & 0.56 & 0.56 & 0.61 & 0.92 & --- \\
\midrule
\rowcolor{fahrow} \oursI & \best{0.77} & \best{0.95} & \best{0.77} & 0.74 & \best{0.79} & \best{0.80} & \best{0.98} & 0.84 \\
\bottomrule
\end{tabular}
\end{table*}

\begin{table*}[!tp]
\centering
\caption{\textbf{Additional stress-test metrics.} PCK@0.05 and fingertip PCK@0.1 (tips) complement the results in \cref{tab:m2d}.}
\label{tab:hardlow}
\tabfont
\setlength{\tabcolsep}{8pt}
\begin{tabular}{lrrrrrr}
\toprule
 & \multicolumn{2}{c}{challenging poses} & \multicolumn{2}{c}{low light} & \multicolumn{2}{c}{shadows} \\
\cmidrule(lr){2-3} \cmidrule(lr){4-5} \cmidrule(lr){6-7}
method & PCK@0.05 & tips & PCK@0.05 & tips & PCK@0.05 & tips \\
\midrule
MediaPipe Hands~\cite{mediapipehands} & 0.75 & 0.91 & 0.34 & 0.57 & 0.70 & 0.88 \\
RTMPose-m~\cite{rtmpose} & 0.60 & 0.85 & 0.40 & 0.64 & 0.83 & 0.92 \\
RTMW-l~\cite{rtmw} & 0.72 & 0.76 & 0.36 & 0.57 & 0.71 & 0.83 \\
HaMeR~\cite{hamer} & 0.77 & 0.92 & 0.36 & 0.62 & 0.90 & 0.95 \\
HaMeR + AnyHand~\cite{anyhand} & 0.71 & 0.87 & 0.26 & 0.55 & 0.82 & 0.91 \\
Hamba~\cite{hamba} & 0.79 & 0.92 & 0.37 & 0.62 & 0.89 & 0.94 \\
WiLoR~\cite{wilor} & 0.79 & 0.92 & 0.34 & 0.62 & 0.91 & 0.94 \\
WiLoR + AnyHand~\cite{anyhand} & 0.82 & 0.92 & 0.32 & 0.59 & 0.88 & 0.94 \\
\rowcolor{fahrow} \oursI & \best{0.92} & \best{0.98} & \best{0.56} & \best{0.74} & \best{0.95} & \best{0.98} \\
\bottomrule
\end{tabular}
\end{table*}

\paragraph{Accuracy and coverage (\cref{tab:hot3d,tab:bench2d,tab:hardlow}).}
AnyHand checkpoints give HOT3D PA-MPJPE of 8.19\,mm for HaMeR and 6.70\,mm for WiLoR. Their per-dataset and stress-test results are included here; the original checkpoints' primary scores are in \cref{tab:m3d,tab:m2d}.

\paragraph{Data-mix transfer (\cref{tab:mix}).}
\label{sec:supp_mix}
More third-person data improves public 3D and challenging-pose results but hurts egocentric tracking-label benchmarks. Increasing FreiHAND weight drives this trade-off; hard-example and HaGRID additions alone do not. We therefore retain the original mixture and use held-out EgoDex part~2 for egocentric selection.

\begin{table}[!htbp]
\centering
\caption{\textbf{Third-person data against tracking ground truth} (ViT-B development setup; PA-MPJPE in mm; hard poses: PCK@0.05). EgoStandard exposure: \cref{sec:m_evalscope}.}
\label{tab:mix}
\tabfont
\setlength{\tabcolsep}{1.5pt}
\begin{tabular}{lrrrrr}
\toprule
 & Dex- & & hard & EgoDex & Ego- \\
model & YCB & HOT3D & poses & p.~2 & Std. \\
\midrule
original mix & 5.68 & 4.41 & 0.91 & \best{6.27} & \best{7.08} \\
+ 3rd person & \best{5.61} & \best{4.39} & \best{0.92} & 6.99 & 7.42 \\
\bottomrule
\end{tabular}
\end{table}

The released pose models omit MOW, ARCTIC and HOI4D pose labels; these datasets are used for contact supervision as described in \cref{sec:supp_chead}.

\paragraph{Released models.}
\label{sec:supp_capacity}
The encoders and input sizes are defined in \cref{sec:m_pose}. All initialize from the ViT-B development heads and share two 12k-step blocks followed by 6k steps with WiLoR targets, at learning rates $10^{-4}$ / $5\cdot10^{-5}$ / $3\cdot10^{-5}$. The part head then trains on frozen features.

\paragraph{Backbone and resolution (\cref{tab:supp_res}).}
DexYCB gains from higher resolution are smaller than HOT3D gains. The supplementary comparison adds mesh errors and stress-test scores to \cref{tab:mres}.

\begin{table}[!htbp]
\centering
\caption{\textbf{Additional backbone and resolution metrics.} The configurations are those of \cref{tab:mres}. Errors are in mm; challenging-pose and low-light scores use PCK@0.1.}
\label{tab:supp_res}
\tabfont
\setlength{\tabcolsep}{2pt}
\begin{tabular}{llrrrrr}
\toprule
 & & HOT3D & \multicolumn{2}{c}{DexYCB} & hard & dark \\
\cmidrule(lr){4-5}
enc. & px & PVE & PA & PVE & PCK & PCK \\
\midrule
ViT-S & 256 & 5.09 & 6.45 & 6.11 & 0.98 & 0.76 \\
 & 320 & 4.99 & 6.42 & 6.09 & 0.98 & 0.76 \\
 & 448 & 4.94 & 6.39 & 6.06 & 0.98 & 0.76 \\
\midrule
ViT-B & 256 & 4.45 & 5.84 & 5.52 & \best{0.99} & 0.79 \\
 & 320 & 4.38 & 5.82 & 5.51 & \best{0.99} & 0.79 \\
 & 448 & 4.34 & 5.79 & 5.47 & \best{0.99} & 0.79 \\
\midrule
ViT-L & 256 & 4.17 & 5.55 & 5.25 & \best{0.99} & \best{0.81} \\
 & 320 & 4.11 & 5.54 & 5.24 & \best{0.99} & \best{0.81} \\
 & 448 & 4.06 & 5.52 & 5.22 & \best{0.99} & 0.80 \\
\midrule
ViT-H+ & 448 & \best{3.99} & \best{5.33} & \best{5.05} & \best{0.99} & 0.80 \\
\bottomrule
\end{tabular}
\end{table}

\begin{table*}[!tp]
\centering
\caption{\textbf{Image-model milestones.} Each row adds to the preceding configuration (\cref{sec:supp_path}). Errors are in mm; 2D scores use PCK@0.1. Top rows use YOLOv8m-pose at 800\,px, except mined and EgoStandard evaluations, which use the one-stage detector. Bottom rows use the one-stage detector throughout. Dense-row PVE evaluates the raw surface. Timing uses CUDA graphs; $^\dagger$marks compiled 384\,px inference.}
\label{tab:abl_image}
\tabfont
\setlength{\tabcolsep}{2.5pt}
\begin{tabular}{lrrrrrrrrrrrr}
\toprule
 & \multicolumn{4}{c}{HOT3D v1} & \multicolumn{2}{c}{bench2d v2} & hard & dark & mined & \multicolumn{2}{c}{EgoStd.} & \\
\cmidrule(lr){2-5} \cmidrule(lr){6-7} \cmidrule(lr){11-12}
milestone & PA & PA-p & PVE & root & PCK & 3D fit & PCK & PCK & PCK & PA & PVE & fps \\
\midrule
\multicolumn{13}{l}{\emph{pose network}} \\
SAM~3D Body hand, off the shelf & 8.30 & --- & 8.49 & 40.00 & 0.61 & --- & --- & --- & --- & --- & --- & 10$^\S$ \\
fine-tuned hand-only teacher (DINOv3-H+) & 4.97 & --- & 5.67 & 11.25 & 0.81 & 0.84 & --- & --- & --- & --- & --- & 18$^\dagger$ \\
ViT-B student, MHR decoder and keypoint head & 4.85 & 5.44 & 5.70 & 11.59 & 0.83 & \best{0.85} & \best{1.00} & 0.80 & --- & 7.59 & 8.09 & 31 \\
decoder-free dense head and root depth & 4.79 & 5.37 & 5.69 & 9.81 & \best{0.84} & 0.76 & 0.99 & \best{0.82} & --- & --- & --- & 54 \\
+ meshes and masks & 4.57 & 5.16 & 5.62 & 8.84 & \best{0.84} & --- & 0.98 & \best{0.82} & --- & 8.35 & 10.66 & --- \\
+ reprojection consistency & 4.38 & 4.96 & 5.59 & \best{8.67} & \best{0.84} & 0.82 & 0.98 & \best{0.82} & 0.47 & 8.15 & 10.70 & --- \\
+ label conventions & 4.53 & 5.11 & 5.80 & 8.79 & \best{0.84} & 0.82 & 0.99 & \best{0.82} & --- & --- & --- & --- \\
+ 400 surface points and data & 4.40 & 4.99 & 5.66 & 8.73 & \best{0.84} & 0.82 & 0.99 & \best{0.82} & --- & --- & --- & --- \\
\midrule
\multicolumn{13}{l}{\emph{image model}} \\
+ one-stage detector, verification, crops & 4.40 & 4.67 & 5.66 & 9.08 & 0.83 & 0.82 & 0.99 & 0.80 & \best{0.98} & \best{7.08} & 9.08 & 46 \\
+ learned MANO fit & 4.40 & 4.68 & 4.08 & 9.07 & 0.83 & 0.82 & 0.99 & 0.80 & \best{0.98} & \best{7.08} & 7.28 & 46 \\
+ fast inference & 4.41 & 4.65 & 4.08 & 9.10 & 0.83 & 0.82 & 0.99 & 0.81 & \best{0.98} & \best{7.08} & 7.28 & \best{85} \\
\rowcolor{fahrow} + ViT-L student and fingertips: \oursI & \best{4.31} & \best{4.58} & \best{4.02} & 8.89 & 0.83 & 0.83 & 0.99 & 0.81 & \best{0.98} & 7.10 & \best{7.12} & 65 \\
\bottomrule
\end{tabular}
\end{table*}

\paragraph{Image milestones (\cref{tab:abl_image}).}
Mesh and mask supervision, reprojection consistency and label harmonization improve complementary aspects of accuracy. The learned MANO fit provides the largest mesh gain; runtime changes nearly double throughput without changing accuracy.

\paragraph{Hard in-the-wild examples.}
Detection, re-cropping and mask-derived boxes account for most gains on this failure-mined set. Person-disjoint HaGRID gestures still share pose patterns with training, so these results are a stress test rather than neutral generalization evidence. Edge-on fingers remain difficult.

\begin{table}[!htbp]
\centering
\caption{\textbf{Detector milestones.} Recall with an early dense pose model, verification and 800\,px input; s/m: small/medium, pose/seg: keypoint/segmentation detector. Mask-bone boxes combine masks and keypoint bones (\cref{sec:supp_det}). fps measures box-only detection on 1408$^2$ frames; WiLoR uses 640\,px.}
\label{tab:det}
\tabfont
\setlength{\tabcolsep}{3pt}
\begin{tabular}{lrrrrr}
\toprule
detector & fps & HOT3D & 2D & hard & dark \\
\midrule
WiLoR's v8m-pose @640 & 123 & 0.96 & 0.92 & 0.84 & 0.81 \\
v8m-pose + verification & 100 & \best{1.00} & 0.95 & 0.93 & 0.89 \\
v8s-pose, low light & \best{158} & \best{1.00} & \best{0.97} & 0.93 & \best{0.95} \\
v8s-seg & 143 & \best{1.00} & \best{0.97} & 0.92 & \best{0.95} \\
v8m-seg & 87 & \best{1.00} & \best{0.97} & 0.94 & \best{0.95} \\
v8m-seg, mask-bone boxes & 93 & \best{1.00} & \best{0.97} & \best{0.95} & \best{0.95} \\
\bottomrule
\end{tabular}
\end{table}

\paragraph{Detectors (\cref{tab:det}).}
Retraining, verification and higher-resolution training improve recall. Mask-derived boxes further improve coverage; a higher fused threshold trades recall for precision. Detector settings are detailed in \cref{sec:supp_det}.

\subsection{Video model}
\label{sec:exp_video}

\begin{table*}[!tp]
\centering
\caption{\textbf{Video-model milestones.} Errors are in mm; acceleration and jitter are in mm/frame$^2$. Top rows add components in order, using the image model's hands; MANO fitting begins at the fourth row. Bottom rows replace one design choice at a time. ``Blend 1'' uses fully refined articulation, and ``det.\ 2nd frame'' runs detection on alternate frames. $^p$Results use stored per-frame predictions.}
\label{tab:abl_video}
\tabfont
\setlength{\tabcolsep}{2.5pt}
\begin{tabular}{lrrrrrrrrrrrrr}
\toprule
 & \multicolumn{5}{c}{HOT3D v1} & \multicolumn{4}{c}{EgoStandard} & \multicolumn{3}{c}{DexYCB} & \\
\cmidrule(lr){2-6} \cmidrule(lr){7-10} \cmidrule(lr){11-13}
milestone & PA & PA-p & PVE & accel & jitter & PA & PVE & accel & jitter & PA & PVE & accel & fps \\
\midrule
per-frame hands & 4.40 & 4.67 & 5.66 & 4.13 & 4.46 & \best{7.08} & 9.08 & 4.53 & 4.60 & 5.72 & 6.82 & 6.94 & 46 \\
+ online refiner & 4.25 & 4.43 & 5.67 & 2.37 & 2.71 & 7.10 & 9.08 & 2.28 & 2.31 & 5.58 & 6.82 & 5.07 & --- \\
+ One Euro & 4.25 & 4.45 & 5.64 & 1.73 & 1.95 & 7.12 & 9.08 & 1.81 & 1.77 & 5.59 & 6.81 & 4.38 & 42 \\
+ MANO fit on every frame & 4.25 & 4.46 & 4.00 & 1.72 & 1.95 & 7.12 & 7.30 & 1.81 & 1.77 & 5.60 & 5.39 & 4.43 & 43 \\
+ fast-motion, 2D filter & 4.23 & 4.39 & 3.99 & 1.75 & 2.01 & 7.13 & 7.30 & 1.80 & 1.77 & 5.53 & 5.32 & 4.29 & \best{80} \\
+ ViT-L hands: \mLV & \best{4.16} & \best{4.38} & \best{3.95} & 1.65 & 1.92 & 7.17 & \best{7.15} & \best{1.67} & \best{1.66} & \best{5.29} & \best{5.10} & \best{3.93} & 59 \\
\midrule
earlier-model refiner$^p$ & 4.29 & 4.50 & --- & 2.65 & --- & 7.20 & --- & 2.19 & --- & 5.87 & --- & 4.95 & --- \\
offline refiner$^p$ & 4.33 & 4.51 & --- & 3.85 & --- & \best{7.08} & --- & 4.29 & --- & 5.66 & --- & 6.41 & --- \\
blend 1 & 4.27 & 4.46 & 5.65 & \best{1.63} & \best{1.79} & 7.17 & 9.08 & 1.81 & 1.74 & 5.60 & 6.82 & 4.23 & --- \\
det.\ 2nd frame & 4.27 & 4.77 & 5.69 & 3.06 & 3.37 & 7.12 & 9.09 & 2.91 & 2.95 & 5.57 & 6.80 & 5.65 & --- \\
\bottomrule
\end{tabular}
\end{table*}

\begin{table}[!htbp]
\centering
\caption{\textbf{Video module design} on an early dense ViT-B model (HOT3D v1 PA-MPJPE / PA-MPJPE-p in mm, accel in mm/frame$^2$). 3rd p.: root correction (cm) applied to a downward-pointing hand in a third-person video.}
\label{tab:refiner}
\tabfont
\setlength{\tabcolsep}{2.5pt}
\begin{tabular}{lrrrrrrr}
\toprule
 & \multicolumn{3}{c}{online} & \multicolumn{3}{c}{offline} & \\
\cmidrule(lr){2-4} \cmidrule(lr){5-7}
training & PA & PA-p & accel & PA & PA-p & accel & 3rd p. \\
\midrule
per frame & 4.79 & 5.37 & 4.23 & 4.79 & 5.37 & 4.23 & --- \\
refiner & 4.34 & 4.71 & \best{2.11} & 4.24 & 4.54 & 2.47 & 13--16 \\
+ geometric aug. & \best{4.23} & \best{4.60} & 2.18 & \best{4.14} & \best{4.45} & \best{2.46} & 13.4 \\
+ rotation aug. & 4.35 & 4.74 & 2.22 & --- & --- & --- & \best{1.7} \\
\bottomrule
\end{tabular}
\end{table}

\paragraph{Refiner design (\cref{tab:refiner}).}
Joint and root losses alone can worsen articulation, motivating the Procrustes term and partial blend. Geometric augmentation improves position invariance; in-plane rotation reduces the third-person root-correction bias from 13.4 to 1.7\,cm. The released refiner uses both augmentations.

\subsection{Camera model and hands in world space}
\label{sec:exp_world}

\paragraph{Where the world-space error comes from (\cref{tab:supp_camdiag}).}
Metric scale explains much of the gap to ground-truth cameras; replacing rotations alone does not help. Translation under-scaling in the motion head motivates the parallax gate. These diagnostics separate camera limitations from hand-reconstruction error.

\begin{table}[!htbp]
\centering
\caption{\textbf{Camera error diagnostics} (W / WA, mm; EgoStandard dev, seed~0; an earlier \mL video model supplies hand masks and scored hands). Each row substitutes a ground-truth component into the reference camera; a diagnostic, not a development progression, so no best marks.}
\label{tab:supp_camdiag}
\tabfont
\setlength{\tabcolsep}{2pt}
\begin{tabular}{lrr}
\toprule
camera (EgoStandard dev) & 100-frame chunks & whole tracks \\
\midrule
causal camera, 5 scale calls & 60.30 / 21.81 & 189.38 / 53.58 \\
+ GT rotations & 64.33 / 22.79 & 195.52 / 54.85 \\
+ GT metric scale per clip & 48.27 / 19.47 & 121.71 / 46.36 \\
+ GT scale per 81-frame segment & 42.87 / 18.60 & 108.69 / 38.99 \\
GT cameras & 32.50 / 13.96 & 64.64 / 16.37 \\
\bottomrule
\end{tabular}
\end{table}
\begin{table}[!htbp]
\centering
\caption{\textbf{World-space tracks on EgoStandard.} Run lengths are in frames; common-set W/WA errors are in mm. Protocol: \cref{sec:benchmarks}.}
\label{tab:supp_tracks}
\tabfont
\setlength{\tabcolsep}{2pt}
\begin{tabular}{llrrrr}
\toprule
hands & camera & runs & med. / p90 & top-10 & common \\
\midrule
\rowcolor{fahrow} \mL & DROID-SLAM & 326 & 900 / 900 & 0.55 & 203.00 / 46.13 \\
\rowcolor{fahrow} \mL & \paper-camera & 326 & 900 / 900 & 0.24 & 141.31 / 38.67 \\
\rowcolor{fahrow} \mL & offline & 326 & 900 / 900 & 0.30 & 124.32 / 35.36 \\
HaWoR~\cite{hawor} & DROID-SLAM & 677 & 137 / 900 & 0.45 & 210.46 / 53.59 \\
WiLoR~\cite{wilor} & DROID-SLAM & 461 & 343 / 900 & 0.45 & --- \\
\bottomrule
\end{tabular}
\end{table}
\begin{table*}[!tp]
\centering
\caption{\textbf{Standard deviations of world-space errors over three runs.} Means are in \cref{tab:mworld}. Runs vary camera seeds, SLAM realizations or both; the extra mask-control row uses our hand masks in HaWoR's camera pipeline. Units match the main table.}
\label{tab:supp_worldsd}
\tabfont
\setlength{\tabcolsep}{1.2pt}
\begin{tabular}{llrrrrrrrrrr}
\toprule
 & & \multicolumn{5}{c}{HOT3D v1} & \multicolumn{5}{c}{EgoStandard} \\
\cmidrule(lr){3-7} \cmidrule(lr){8-12}
 & & \multicolumn{2}{c}{chunks} & \multicolumn{2}{c}{whole} & & \multicolumn{2}{c}{chunks} & \multicolumn{2}{c}{whole} & \\
hands & camera & W & WA & W & WA & RTE & W & WA & W & WA & RTE \\
\midrule
WiLoR~\cite{wilor} & DROID-SLAM & 0.19 & 0.05 & 0.25 & 0.07 & 0.02 & 4.26 & 0.23 & 31.75 & 1.00 & 0.06 \\
HaWoR~\cite{hawor} & DROID-SLAM & 0.21 & 0.05 & 0.26 & 0.07 & 0.02 & 4.11 & 0.25 & 28.90 & 1.21 & 0.08 \\
\rowcolor{fahrow} \mLV & DROID-SLAM & 0.21 & 0.07 & 0.41 & 0.12 & 0.02 & 4.79 & 0.23 & 41.97 & 1.08 & 0.06 \\
\rowcolor{fahrow} \mLV & DROID-SLAM, our masks & 0.34 & 0.09 & 0.40 & 0.12 & 0.04 & 1.66 & 0.41 & 18.97 & 1.42 & 0.02 \\
\rowcolor{fahrow} \mLW & \paper-camera & 0.43 & 0.31 & 0.69 & 0.38 & 0.30 & 0.18 & 0.08 & 3.71 & 1.64 & 0.02 \\
\rowcolor{fahrow} \mLW & DROID-SLAM & 0.29 & 0.18 & 0.58 & 0.24 & 0.17 & 0.92 & 0.61 & 3.57 & 1.98 & 0.02 \\
\bottomrule
\end{tabular}
\end{table*}

\paragraph{Track coverage (\cref{tab:supp_tracks}).}
We report median and 90th-percentile run lengths. Top-10 measures the fraction of W error from the ten worst runs. Common-set W/WA averages three runs on hand-frames detected by both \mL and HaWoR. On this shared set, paired 95\% bootstrap intervals for causal and offline gains exclude zero.

\paragraph{Dyn-HaMR.}
Dyn-HaMR~\cite{dynhamr} stages I--II worsened its initializer on HOT3D pilots despite camera corrections. Stage III requires fixed 128-frame tracks, so we exclude it from the variable-length world comparison.

\subsection{Held-out video benchmarks}
\label{sec:exp_heldout}
\label{sec:supp_dev}

\begin{table}[!htbp]
\centering
\caption{\textbf{DexYCB} (s0 test, 160 clips, 9.3k hands; mm, accel and jitter in mm/frame$^2$). $^\dagger$trained on DexYCB; $^\S$with DexYCB training crops; $^\ddagger$refiner also trained on DexYCB training tracks.}
\label{tab:dexycb}
\tabfont
\setlength{\tabcolsep}{1.5pt}
\begin{tabular}{lrrrrrrr}
\toprule
method & recall & PA & PA-p & PVE & accel & jitter & root \\
\midrule
WiLoR$^\dagger$~\cite{wilor} & \best{1.00} & \best{4.87} & \best{5.14} & \best{4.66} & 5.63 & 4.52 & 52.05 \\
\midrule
\rowcolor{fahrow} \oursI$^\S$ & 0.99 & 5.40 & 6.49 & 5.13 & 6.04 & 4.95 & 25.92 \\
\rowcolor{fahrow} \mLV$^{\S\ddagger}$ & 0.99 & 5.29 & 5.91 & 5.10 & \best{3.93} & \best{2.25} & \best{23.87} \\
\bottomrule
\end{tabular}
\end{table}

\begin{table}[!htbp]
\centering
\caption{\textbf{Additional EgoStandard camera-frame metrics.} Root-relative MPJPE and relative hand position (mm), plus detection coverage, complement the primary results in \cref{tab:m3d}.}
\label{tab:heldout}
\tabfont
\setlength{\tabcolsep}{3pt}
\begin{tabular}{lrrrr}
\toprule
method & rr & MRRPE & recall & prec. \\
\midrule
WiLoR~\cite{wilor} & 28.10 & 44.38 & 0.99 & \best{0.98} \\
\midrule
\rowcolor{fahrow} \oursI & \best{17.05} & \best{32.97} & \best{1.00} & \best{0.98} \\
\rowcolor{fahrow} \mLV & 17.15 & 33.12 & \best{1.00} & \best{0.98} \\
\bottomrule
\end{tabular}
\end{table}

\paragraph{DexYCB (\cref{tab:dexycb}).}
WiLoR retains the best articulation and mesh accuracy, while \mLV has the best temporal metrics. Precision is omitted because the second hand is unannotated.

\subsection{Points versus parameters}
\label{sec:exp_dense}

\begin{table}[!htbp]
\centering
\caption{\textbf{Points versus parameters.} We compare WiLoR's MANO output, \mMHR, the dense ViT-B model after reprojection consistency, and the final development model with MANO fitting. HOT3D errors are in mm; bench2d scores are PCK@0.1 for direct 2D and fitted projections. Dense PVE uses the raw surface.}
\label{tab:dense_vs_param}
\tabfont
\setlength{\tabcolsep}{1.6pt}
\begin{tabular}{lrrrrrrr}
\toprule
model & PA & 2D & 3D fit & root & MRRPE & PVE & fps \\
\midrule
WiLoR & 6.72 & 0.69 & 0.77 & 43.89 & 48.57 & 6.49 & 4.7 \\
\rowcolor{fahrow} \mMHR & 4.85 & 0.83 & \best{0.85} & 11.59 & 14.65 & 5.70 & 31 \\
dense ViT-B & \best{4.38} & \best{0.84} & 0.82 & \best{8.67} & \best{11.67} & 5.59 & 53 \\
last milestone + fit & 4.41 & 0.83 & 0.82 & 9.10 & 12.48 & \best{4.08} & \best{85} \\
\bottomrule
\end{tabular}
\end{table}

\paragraph{Against an MHR head (\cref{tab:dense_vs_param}).}
Reprojection consistency narrows the gap between direct 2D and fitted 3D projections. The supplementary comparison also reports root placement, relative hand position and runtime for the development models.

\begin{table}[!htbp]
\centering
\caption{\textbf{Mesh fitting.} PA-MPVPE (mm) on development image predictions. The learned parameter head shares the network of \cref{tab:mdense}; GN: Gauss--Newton; fitter input noise: 3\,mm unless noted. Bold row: default fitting stage.}
\label{tab:meshfit}
\tabfont
\setlength{\tabcolsep}{4pt}
\begin{tabular}{lrr}
\toprule
mesh & HOT3D & EgoStandard \\
\midrule
WiLoR (MANO) & 6.49 & 7.52 \\
\midrule
dense surface & 5.66 & 9.08 \\
learned MANO-parameter head & 6.79 & 12.05 \\
closed form alone & 6.51 & --- \\
closed form + 3 GN steps & 4.49 & 7.42 \\
learned fitter, no input noise & \best{4.03} & 8.40 \\
\textbf{learned fitter + 1 GN step} & 4.08 & \best{7.28} \\
\bottomrule
\end{tabular}
\end{table}

\begin{figure*}[!tp]
\centering
\includegraphics[width=\linewidth]{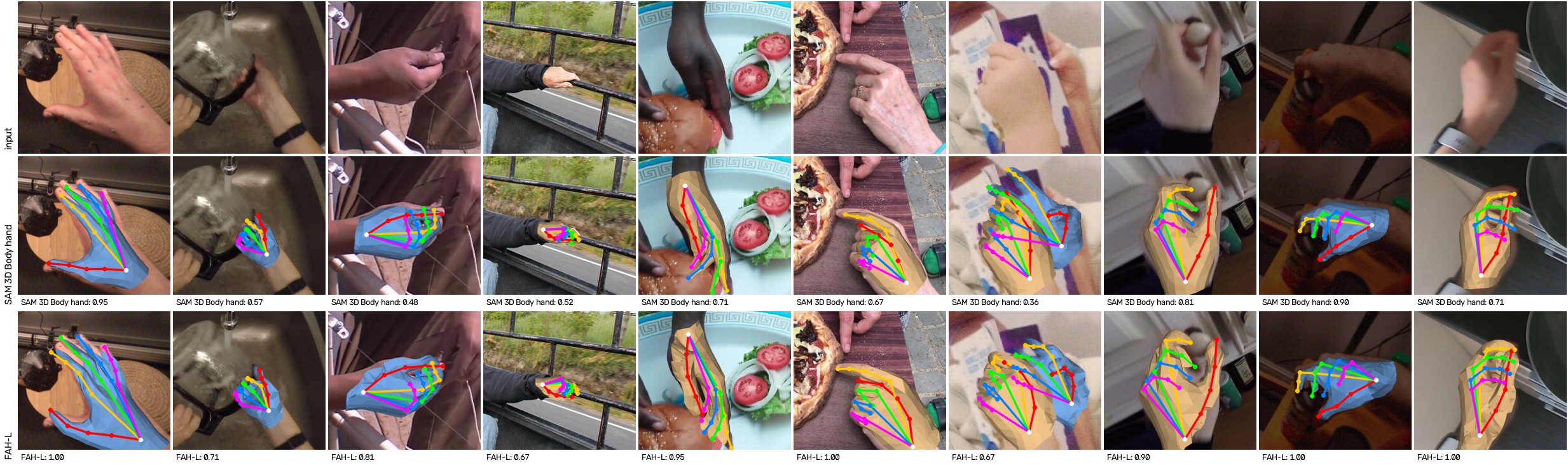}
\caption{\textbf{SAM~3D Body hand branch versus \mL} on bench2d v2 (HOT3D, COCO, HInt; ninth: synthetic low light). Scores: per-hand PCK@0.1, projected HOT3D GT/human labels elsewhere, averaged for two hands. Each stratum shows the first correctly sided hand found by both in fixed random order; outdoor/blur additionally require $\geq$0.1 FAH gain. FAH overlays use dense surfaces without fitting; the timed pipeline includes the fit. Failures: \cref{fig:supp_failures}; sources: \cref{sec:supp_credits}.}
\label{fig:mcompare}
\end{figure*}

\paragraph{Mesh fitting (\cref{tab:meshfit}).}
Input-noise augmentation is essential for transfer to EgoStandard (\cref{tab:meshfit}).

\begin{table*}[!tp]
\centering
\caption{\textbf{Number of dense surface points.} All models share data, initialization and an 8k-step label-convention fine-tune. Reduced-point models reconstruct 778 vertices with $K=N/2$ spectral modes; the 778-point row is unsmoothed. HOT3D errors are in mm, and 2D scores use PCK@0.1. Speed differences are below run-to-run variation.}
\label{tab:npoints}
\tabfont
\setlength{\tabcolsep}{3pt}
\begin{tabular}{lrrrrrrrrrrr}
\toprule
points & PA & PA-p & PVE & PVE-p & F@5 & root & 2D & 3D fit & hard & dark & fps \\
\midrule
778 & 4.40 & 4.99 & 5.66 & 6.21 & \best{0.80} & \best{8.54} & \best{0.84} & \best{0.83} & \best{0.98} & 0.82 & 55 \\
400 & 4.30 & 4.88 & \best{5.40} & \best{5.95} & 0.78 & 8.64 & \best{0.84} & \best{0.83} & \best{0.98} & \best{0.83} & 53 \\
200 & \best{4.27} & \best{4.85} & 5.48 & 6.03 & 0.76 & 8.80 & \best{0.84} & \best{0.83} & \best{0.98} & \best{0.83} & \best{56} \\
\bottomrule
\end{tabular}
\end{table*}

\paragraph{Surface points (\cref{tab:npoints}).}
The 400-point model gives the best mesh accuracy at similar throughput. Fewer points increase the representation floor and lose surface detail.

\paragraph{Agreement with the image (\cref{tab:mdense,tab:magree}).}
\label{sec:supp_align}
Distances are normalized by ground-truth hand size; mesh joints use the MANO regressor and fingertip vertices. Rigid PnP improves joint alignment but worsens surface alignment. Adding surface points to PnP can shrink close-up fingers, so it is not the default.

\begin{table}[!htbp]
\centering
\caption{\textbf{Fingertip extent on bench2d v2 matched hands.} Signed tip error is measured along the labeled distal segment in segment lengths; negative values indicate shortfall. Mesh extent uses the furthest tip-region vertex. Length is the median mesh-to-labeled finger-length ratio. Ideal values are zero error and unit length.}
\label{tab:supp_tips}
\tabfont
\setlength{\tabcolsep}{3pt}
\begin{tabular}{lrrrr}
\toprule
 & joints & mesh & HOT3D & length \\
\midrule
SAM~3D Body hand & $-$0.18 & $-$0.14 & $-$0.19 & 0.81 \\
\rowcolor{fahrow} \mL, no tip step & $-$0.06 & $+$0.01 & 0.00 & 0.99 \\
\rowcolor{fahrow} \mL & $-$0.05 & $+$0.02 & $+$0.02 & 1.00 \\
WiLoR~\cite{wilor} & $+$0.05 & $+$0.12 & $+$0.47 & 0.97 \\
\bottomrule
\end{tabular}
\end{table}

\paragraph{Fingertips (\cref{tab:supp_tips}).}
The fitted MANO mesh has little median fingertip shortfall. The mask-guided step improves direct 2D and dense-surface extent at a small runtime and HOT3D mesh-accuracy cost, with little change on EgoStandard. It is independent of the part head and applies only to distal segments spanning at least six mask cells.

\par\medskip

\subsection{Detection and instance masks in one stage}
\label{sec:exp_segdet}

The detector masks cover the hand rather than the forearm; \cref{tab:masks} compares the released pipeline with segmentation baselines.

\begin{table}[!htbp]
\centering
\caption{\textbf{VISOR validation segmentation.} We report mIoU for hand-and-arm (h+a) and wrist-cut hand (h) masks, and inference speed. $^o$Ground-truth box oracle. VISOR-HOS is trained on VISOR.}
\label{tab:masks}
\tabfont
\setlength{\tabcolsep}{2.5pt}
\begin{tabular}{lrrr}
\toprule
 & \multicolumn{2}{c}{VISOR} & \\
\cmidrule(lr){2-3}
method & h+a & h & fps \\
\midrule
\rowcolor{fahrow} \mL & 0.54 & 0.78 & 54 \\
\paper detector & 0.53 & 0.77 & \best{253} \\
SAM~2.1-B~\cite{sam2}, boxes & 0.55 & 0.77 & 23 \\
SAM~2.1-L, boxes & 0.56 & 0.78 & 15 \\
SAM~3~\cite{sam3}, boxes & 0.56 & \best{0.81} & 8.4 \\
SAM~3, text & 0.59 & \best{0.81} & 6.3 \\
VISOR-HOS~\cite{visor} & \best{0.90} & 0.79 & 21 \\
Hands23~\cite{hands23} & 0.50 & 0.72 & 11 \\
\midrule
SAM~2.1-L, GT boxes$^o$ & 0.94 & 0.83 & --- \\
\bottomrule
\end{tabular}
\end{table}

\paragraph{Timing and prompts.}
We time one of four shards from 3,682 decoded 1920$\times$1080 frames. Synchronized mask calls include image encoding and full-resolution output, but exclude loading and three warm-up calls. We compute fps from the reciprocal of the median of chunk-median runtimes.

SAM~2.1/3 use eager PyTorch in bf16; VISOR-HOS and Hands23 use their full Detectron2 pipelines in fp32. SAM box prompts come from the released detector, whose runtime is excluded; the text prompt is ``hand.'' For \mL, timing includes pose estimation and mask export on every frame. The detector alone uses TensorRT FP16 at 640\,px.

\paragraph{Against segmentation models (\cref{tab:masks}).}
The pipeline exports masks with pose faster than the segmentation baselines, but most achieve higher VISOR accuracy. VISOR-HOS also benefits from training on VISOR and predicting the forearm.

\subsection{Contact}
\label{sec:supp_contact}

HOI4D evaluation excludes Chair and holds out subject H3, giving 7,280 crops from 149 sequences. ARCTIC validation contains only 612 hands from four sequences, so uncertainty is high. MOW uses a 35\,mm tolerance and HACO threshold 0.52.

\paragraph{Qualitative selection (\cref{fig:mcontact}).}
Examples use ground-truth boxes and the same ground-truth MANO mesh, viewed from the contact side; contact labels use the 10\,mm rule of \cref{tab:mcontact}. From a fixed random order, we select the first hand with at least 30 contact vertices for each grasp type (HOT3D pinch, power, fingers; DexYCB pinch, power), without repeating objects, followed by the first \mL failure with vertex F1 below 0.3.

\begin{table*}[!tp]
\centering
\caption{\textbf{Additional contact metrics.} HOT3D and DexYCB use mean per-hand vertex F1; HOI4D and ARCTIC also report pooled F1/AP, and MOW uses per-image F1. Primary HOT3D/DexYCB results are in \cref{tab:mcontact}. HACO is excluded from best-value marks.}
\label{tab:supp_cfamily}
\tabfont
\setlength{\tabcolsep}{2.4pt}
\begin{tabular}{lrrrrrrrrr}
\toprule
 & HOT3D & DexYCB & \multicolumn{3}{c}{HOI4D (no Chair)} & \multicolumn{3}{c}{ARCTIC val} & MOW \\
\cmidrule(lr){2-3} \cmidrule(lr){4-6} \cmidrule(lr){7-9}
model & hand & hand & F1 & AP & hand & F1 & AP & hand & F1 \\
\midrule
HACO~\cite{haco} & 0.60 & 0.66 & --- & --- & --- & --- & --- & --- & 0.52 \\
\rowcolor{fahrow} \mS & 0.74 & 0.69 & 0.62 & 0.61 & 0.54 & 0.74 & 0.79 & 0.69 & \best{0.58} \\
\rowcolor{fahrow} \mBrel & 0.76 & 0.71 & 0.63 & 0.62 & \best{0.56} & 0.76 & 0.80 & 0.70 & 0.57 \\
\rowcolor{fahrow} \mL & 0.76 & 0.71 & \best{0.64} & \best{0.64} & \best{0.56} & 0.76 & \best{0.81} & \best{0.72} & 0.56 \\
\rowcolor{fahrow} \mH & \best{0.77} & \best{0.72} & 0.63 & 0.62 & 0.55 & \best{0.77} & \best{0.81} & \best{0.72} & 0.53 \\
\bottomrule
\end{tabular}
\end{table*}

\begin{figure*}[t]
  \centering
  \includegraphics[width=\linewidth]{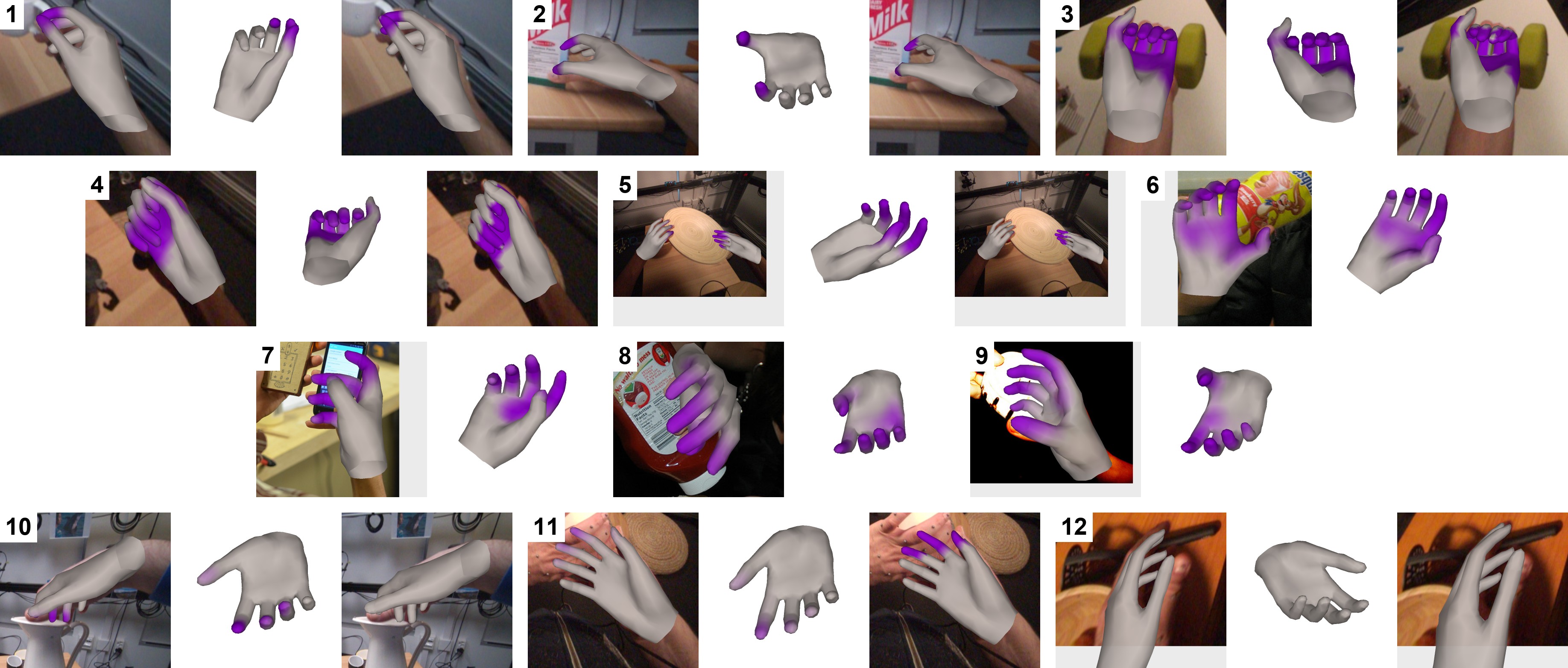}
  \caption{\textbf{Selected \mL contact examples.} Panels 1--5 and 10--12 are from training-disjoint HOT3D; 6--9 are from COCO. Each shows fitted-mesh contact, a palm view and ground truth where available. Gray to violet indicates increasing probability, with a calibrated threshold of 0.5. Failures include false contact above a vase (10), missed occluded fingertip contact (11) and missed stick contact without a ground-truth object pose (12). Credits are in \cref{sec:supp_credits}.}
  \label{fig:supp_contact}
\end{figure*}
\begin{table}[!htbp]
\centering
\caption{\textbf{Contact data ablation.} Two-seed means for \mL; vertex F1/AP and MOW per-image F1 (HACO protocol). All recipes include HOT3D, DexYCB and MOW. Added ARCTIC/HOI4D data make those evaluations in-domain; HOI4D excludes Chair and holds out the test subject.}
\label{tab:supp_cmix}
\tabfont
\setlength{\tabcolsep}{1.5pt}
\begin{tabular}{lrrrrrrrrr}
\toprule
 & \multicolumn{2}{c}{HOT3D} & \multicolumn{2}{c}{DexYCB} & MOW & \multicolumn{2}{c}{HOI4D} & \multicolumn{2}{c}{ARCTIC} \\
\cmidrule(lr){2-3} \cmidrule(lr){4-5} \cmidrule(lr){7-8} \cmidrule(lr){9-10}
data & F1 & AP & F1 & AP & F1 & F1 & AP & F1 & AP \\
\midrule
original & 0.80 & 0.85 & 0.76 & 0.81 & 0.51 & 0.52 & 0.52 & 0.68 & 0.70 \\
+ ARCTIC & 0.80 & 0.85 & 0.76 & 0.82 & 0.53 & 0.54 & 0.53 & 0.76 & 0.81 \\
+ ARCTIC, HOI4D & 0.80 & 0.85 & 0.76 & 0.82 & 0.54 & 0.64 & 0.63 & 0.76 & 0.81 \\
\bottomrule
\end{tabular}
\end{table}

\paragraph{Training data (\cref{tab:supp_cmix}).}
Adding ARCTIC and HOI4D improves their held-out domains and MOW without a resolved pooled vertex-F1 change on HOT3D or DexYCB. These are in-domain training gains, not zero-shot transfer.

\begin{figure*}[!tp]
  \centering
  \includegraphics[width=\linewidth]{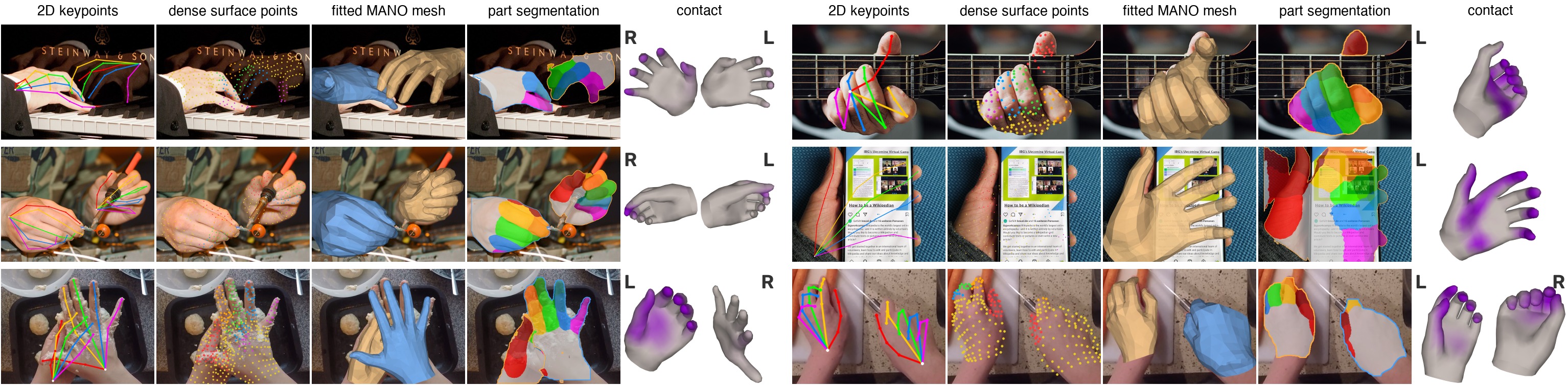}
  \caption{\textbf{\mH failures}, with the five outputs of \cref{fig:teaser}. Piano: flattened left mesh/missed key contact; guitar: tangled fretting-hand joints; soldering: coarse meshes; phone: parts spill onto the screen and label occluded fingers; dough: missed right-hand contact; knife: coarse mesh. Sources: Wikimedia Commons and, bottom row, EPIC-KITCHENS-100 via HInt; public test splits, full credits in \cref{sec:supp_credits}.}
  \label{fig:supp_failures}
\end{figure*}

\section{Image credits}
\label{sec:supp_credits}
{\footnotesize
Wikimedia Commons photos, left to right and top to bottom. \Cref{fig:teaser}: keyboard, Cangjie6 (``Keyboard with fingers'', CC~BY-SA~4.0); knitting, Sadarama (``Oma strickt Str\"umpfe'', CC~BY-SA~4.0); pottery, Earthen Potter (``Earthen Potter Hands'', CC~BY-SA~4.0); Lego, Nenad Stojkovic (``Playing with Lego bricks'', CC~BY~2.0); Rubik's cube, theilr (``Rubik cube (345056969)'', CC~BY-SA~2.0). \Cref{fig:supp_failures}: piano, Quincena Musical (``Mikhail Shehtman (hands playing a piano)'', CC~BY~2.0); guitar, Nenad Stojkovic (``Eko Rio Bravo 12 -- A-mol chord'', CC~BY~2.0); soldering, U.S. Air Force photo by Airman Jessica Struble (``Desoldering'', public domain); smartphone, Mictlancihuatl (``Hand holding Smartphone'', CC~BY-SA~4.0).
The overlays are ours; the other frames come from our held-out HOT3D~\cite{hot3d} benchmark clips (HOT3D's train\_aria recordings, disjoint from our training; \cref{fig:teaser,fig:pipeline,fig:mcontact}; CC~BY-SA~4.0, hand annotations CC~BY-NC-SA~4.0, which give the ground-truth meshes and contact in \cref{fig:mcontact}), the DexYCB~\cite{dexycb} s0 test split (\cref{fig:mcontact}; CC~BY-NC~4.0) and the HInt~\cite{hamer} test split (\cref{fig:teaser,fig:supp_failures}; EPIC-KITCHENS-100 frames~\cite{damen2022rescaling} via HInt, CC~BY-NC~4.0). \Cref{fig:mcompare} (bench2d v2 test images), left to right: 1 and 9 frames of the held-out HOT3D~\cite{hot3d} benchmark clips (CC~BY-SA~4.0, hand annotations CC~BY-NC-SA~4.0; 9 with our synthetic low light); 2, 8 and 10 EPIC-KITCHENS-100~\cite{damen2022rescaling} frames via HInt~\cite{hamer} (CC~BY-NC~4.0); 3, 5 and 6 New Days of Hands~\cite{hands23} frames via HInt (public test split; no separate frame license stated); 4 ``Luggage and all'' by Sally Taylor (Flickr, CC~BY-NC~2.0) and 7 Flickr photo 7173995615 (CC~BY-NC~2.0; author page no longer online), both via COCO val2017 and COCO-WholeBody~\cite{cocowholebody}.
\Cref{fig:supp_contact}: frames of the held-out HOT3D~\cite{hot3d} benchmark clips (CC~BY-SA~4.0, hand annotations CC~BY-NC-SA~4.0) (panels 1--5, 10--12) and COCO train2017 Flickr photos (panels 6--9, in order): ``Snack Break'' by Eden, Janine and Jim (CC~BY~2.0), ``Nadya and Jeff making cellphones'' by David Mellis (CC~BY~2.0), ``vincent's last night in town'' by brooke (CC~BY-SA~2.0) and ``La Creaci\'on del Donut\textregistered'' by ..::Fluckr You::.. (CC~BY-SA~2.0); all cropped, with our renders.
\Cref{fig:supp_samples}, Flickr authors in panel order: Maggie Brauer, Zepfanman.com, Nagarjun Kandukuru, super awesome, Maryland GovPics, rachaelvoorhees, Becky Wetherington and Brad Greenlee. Photo titles: 1 ``Delta\_100\_Ring\_Park073'', 2 ``Zebra hand'', 3 ``Kinu as Capsicum'', 4 ``bratmobile'' (CC~BY~2.0); 5 ``20111005-first lady host wine tasting for nga staff-jb'', 6 ``wii bowling'', 7 ``68/365 -- I'm too sexy to match.'', 8 ``Jelly Doughnut Incident'' (CC~BY~2.0); 1, 3 and 4 via Open Images V7, 2 via Openverse, 5--8 via COCO val2017; all cropped, 5--8 darkened by our low-light synthesis.\par}

\end{document}